\documentclass[10pt,journal,compsoc]{IEEEtran}
\ifCLASSOPTIONcompsoc
  \usepackage[nocompress]{cite}
\else
  \usepackage{cite}
\fi
\ifCLASSINFOpdf
\else
\fi
\usepackage{amsmath}

\usepackage{multirow}
\usepackage{arydshln}
\usepackage{subfigure}
\usepackage{graphicx}
\usepackage{booktabs}  
\usepackage{times}
\usepackage{latexsym}
\usepackage{amsfonts}
\usepackage{amssymb}
\usepackage{hyperref}
\usepackage{svg}
\begin{document}
%
\title{Attributed Abnormality Graph Embedding for Clinically Accurate X-Ray Report Generation}
%
%
%
%

\author{Sixing Yan,
 William K. Cheung,
 Keith Chiu,
 Terence M. Tong,
 Charles K. Cheung,
 Simon See
\IEEEcompsocitemizethanks{\IEEEcompsocthanksitem Sixing Yan and William K. Cheung are with the Department of Computer Science, Hong Kong Baptist University. E-mail: \{cssxyan, william\}@comp.hkbu.edu.hk. \protect\\
Keith Chiu is with the Queen Elizabeth and Kwong Wah Hospitals. Email: kwhchiu@hku.hk \protect\\
Terence Tong is with the Tuen Mun Hospital. Email: tmc877@ha.org.hk \protect\\
Sixing Yan, Charles Cheung and Simon See are with NVIDIA AI Technology Center, NVIDIA Corporation, Hong Kong. Email: \{alfonsoy,chcheung,ssee\}@nvidia.com \protect\\
\IEEEcompsocthanksitem Corresponding author: William K. Cheung. \protect\\
\IEEEcompsocthanksitem This work has been submitted to the IEEE for possible publication. Copyright may be transferred without notice, after which this version may no longer be accessible.
}
}

%
%

\markboth{Journal 2022}%
{Yan \MakeLowercase{\textit{et al.}}: Attributed Abnormality Graph Embedding for Clinically Accurate X-Ray Report Generation}
%



\IEEEtitleabstractindextext{%
\begin{abstract}
Automatic generation of medical reports from X-ray images can assist radiologists to perform the time-consuming and yet important reporting task.
Yet, achieving clinically accurate generated reports remains challenging.
Modeling the underlying abnormalities using the knowledge graph approach has been found promising in enhancing the clinical accuracy.
In this paper, we introduce a novel fined-grained knowledge graph structure called an attributed abnormality graph (ATAG). The ATAG consists of interconnected abnormality nodes and attribute nodes, allowing it to better capture the abnormality details. In contrast to the existing methods where the abnormality graph was constructed manually, we propose a methodology to automatically construct the fine-grained graph structure based on annotations, medical reports in X-ray datasets, and the RadLex radiology lexicon. We then learn the ATAG embedding using a deep model with an encoder-decoder architecture for the report generation. In particular, graph attention networks are explored to encode the relationships among the abnormalities and their attributes. A gating mechanism is adopted and integrated with various decoders for the generation. We carry out extensive experiments based on the benchmark datasets, and show that the proposed ATAG-based deep model outperforms the SOTA methods by a large margin and can improve the clinical accuracy of the generated reports.
\end{abstract}

}

\maketitle

\IEEEdisplaynontitleabstractindextext

%
\IEEEpeerreviewmaketitle

\IEEEraisesectionheading{\section{Introduction}\label{sec:introduction}}

%
%
%
%

 

\begin{figure*}[h]
\centering
\includegraphics[width=0.95\textwidth]{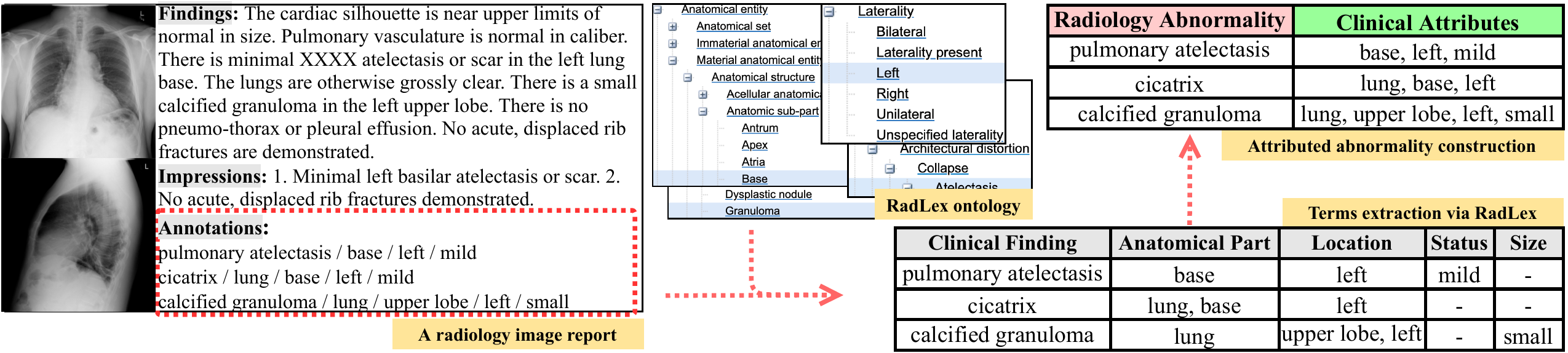}
\caption{Illustration of extracting abnormalities and attributes from the annotated radiology report and the RadLex ontology.} 
\label{fig:example}
\end{figure*}

\IEEEPARstart{A}{utomatic} generation of medical reports from X-ray images has recently been studied with the objective to assist radiologists to perform the time-consuming and yet important reporting task. An X-ray report, as shown in Fig.~\ref{fig:example}, typically contains a paragraph with multiple sentences describing the abnormalities identified by the radiologist in the images (called \textit{findings}) and a short conclusion (called \textit{impression}). For the generated report to be clinically accurate, findings of abnormalities revealed in the X-ray images should be correctly reported.
In the literature, the deep encoder-decoder architecture has been found effective for the medical report generation, where visual features were extracted from the input medical images using a convolutional neural network (CNN) and fed to a recurrent neural network (RNN) to generate the report~ \cite{xue2018multimodal,han2018towards,xie2019attention,yang2021writing}. 
Some recent work~\cite{chen2020generating,miura2021improving,liu2021exploring} replaces the decoder with a Transformer-based architecture for further text quality enhancement.
Other than improving the fluency and readability text of generated report, some study also attempt to increase the clinical keywords accuracy using reinforcement learning~\cite{liu2019clinically,jing2019show,li2018hybrid}, 
For enhancing the clinical accuracy, semantic annotations~\cite{jing2018automatic} (see also  Fig.~\ref{fig:example}) and concepts extracted from the medical reports~\cite{yuan2019automatic} have been used to learn semantic features to assist the report generation.

Recently, the knowledge graph approach integrated into a deep model architecture~ \cite{li2019knowledge,zhang2020when,liu2021exploring, liu2021auto} has been shown effective to further enhance the accuracy.
Among the existing knowledge-graph based medical report generation methods, it is common that careful manual effort is required to construct the abnormality graph. It is inevitable to result in a sub-optimal design. In addition, we notice that a medical report typically contain not only information about the observed abnormalities (e.g., ``calcified granuloma''), but also their associated ``attributes'' (e.g., ``left upper lobe'' as its location). Therefore, it becomes important to represent well both the abnormalities and their attributes in order the generated reports can recover the related details. However, related research work where attributed abnormalities are explicitly represented is still rare. Our conjecture is that constructing a knowledge graph with higher granularity of abnormalities and attributes is vital for enabling the generation of clinically accurate reports.
Orthogonal to this direction, retrieving relevant clinical templates for rewriting is another trick to ease the report generation task~ \cite{li2018hybrid,biswal2020clinical,li2019knowledge,syeda2020chest,liu2021exploring}. 

In this paper, we focus on investigating how the knowledge graph approach can be better exploited for medical report generation. We first propose the adoption of a novel fine-grained knowledge graph structure to represent the attributed abnormalities. To attain such a fine-grained \textbf{AT}tributed \textbf{A}bnormality \textbf{G}raph (ATAG), we propose a methodology to automatically construct it based on annotated X-ray datasets and the RadLex radiology lexicon. In ATAG, each attributed abnormality is represented using an abnormality node and an associated set of attribute nodes to model the abnormality details. The inter-related abnormalities and attributes are connected.
The ATAG 
forms a global medical knowledge graph, and is then integrated into a deep encoder-decoder model architecture to learn its embedding with the objective to achieve highly accurate abnormality classification and high quality radiology report generation.
In addition, a novel gate mechanism is designed to allow the information encoded by ATAG more effectively incorporated into both LSTM- and Transformer-based decoder for clinically accurate report generation. 
Our experimental results based on the publicly available IU-XRay dataset \cite{demner2016preparing} and MIMIC-CXR dataset~\cite{johnson2019mimiccxr} show that the use of ATAG can achieve a higher accuracy on abnormality classification. Also, it can generate more clinically accurate reports compared to the SOTA methods according to the natural language generation metric scores and the medical report quality metrics by a large margin. 
To summarize, 
the main contributions of this paper include:  
\begin{enumerate}
\item A methodology to automatically construct a novel fine-grained attributed abnormality graph (ATAG) representing the abnormalities and their associated attributes;
\item An algorithm to learn the attributed abnormality representations of ATAG using
graph attention networks;
\item A gating mechanism to effectively integrate the ATAG with various decoders to generate the detailed radiology report with clinically accurate attributed abnormalities. 
\end{enumerate}

\section{Related Work}

The earliest efforts on generating textual output based on visual input are for automatic image captioning~\cite{Xu2015Show,vinyals2015show,you2016image,rennie2017self,anderson2018bottom}. The image captioning task typically aims at generating one sentence to describe the objects, their attributes, and the underlying scenes revealed in the given image. Image paraphrasing~\cite{krause2017hierarchical} is a related task which focuses on generating multiple sentences (i.e., paragraphs), instead of just one sentence. The objective is to provide more detailed object-related descriptions or a long sentence with details about the main objects in the image.

\subsection{Radiology report generation}
Generating radiology reports, in a sense, is similar to the image paraphrasing task which takes an X-ray image as input and output a text with multiple sentences. Each sentence usually focus on one particular topic, i.e., a clinical observation, with some fine-grained supporting details revealed in the input medical image. In the literature, deep learning based methods using CNN encoder and RNN decoder for the report generation have been found promising~ \cite{xue2018multimodal,han2018towards,xie2019attention,yang2021writing}. 

\subsection{Use of semantic features/labels}
To achieve the clinically accurate report generation, many recent work that makes use of extra labels under the deep encoder-decoder framework for generating clinically accurate reports. Yuan {\it et al.}~\cite{yuan2019automatic} extracted 69 medical concepts from the medical reports using Semrep~(\texttt{https://semrep.nlm.nih.gov/}), and trained a CNN for concept classification and report generation. Jing {\it et al.}~\cite{jing2018automatic} generated 572 tags for the whole dataset using Medical Text Indexer (MTI), and generated reports using both the semantic features from the tags and the visual features. 
Alternatively, Park \textit{et al.}~\cite{park2020feature} utilized 210 tags and multi-level visual features to facilitate the report generation.
Siddharth \textit{et al.} \cite{biswal2020clinical} learned to predict 14 disease labels based on the visual features 
and fed them to the report generation decoder. Syeda-Mahmood \textit{et al.}~\cite{syeda2020chest} made use of more fine-grained labels (with 78 unique abnormalities and 9 attributes) generated from the reports. 
Miura \textit{et al.}~\cite{miura2021improving} first learned an image-to-text model, and then fine-tuned it by increasing the number of the matched clinical entities in the generated report.
All the aforementioned methods reply on either some taggers or manual effort to select the concepts and the abnormalities.

\subsection{Use of knowledge graphs}
There also exist methods proposed to organize the medical concepts and abnormalities using a knowledge graph. Li \textit{et al.}~\cite{li2019knowledge}
proposed the use of an abnormality-and-attributes graph where the nodes correspond to 80 abnormality phrases (manually chosen) frequently appearing in reports and the edges are constructed according to their occurrence frequencies. A graph Transformer is proposed to dynamically transform the image-to-graph features and graph-to-text features with the attention mechanism.
Zhang \textit{et al.}~\cite{zhang2020when} manually constructed a knowledge graph with 20 common abnormalities where the nodes are connected according to the body parts they appeared. 
Liu \textit{et al.}~\cite{liu2021exploring} further integrated this 20-abnormalities knowledge graph and report templates as the prior knowledge for report generation. 
In this paper, we propose a methodology to automatically construct an abnormality graph with fine-grained attributes to be integrated into a deep learning model for generating reports with a higher clinical accuracy. 

\subsection{Aligning visual features and report contents}
Some more recent research efforts try to better align the abnormal observations and the report contents. For instances,  
Liu \textit{et al.}~\cite{liu2021contrastive} proposed a contrastive attention mechanism to subtract the  ``normality'' visual features from the overall visual features for the decoder to generate the depiction of observed abnormalities. 
You \textit{et al.}~\cite{you2021alignTransformer}  developed an alignment-enhanced Transformer to refine the visual features with the semantic features of disease labels by an alternative alignment mechanism.
The memory mechanism has also been employed for the report generation by memorizing the visual pattern of normal/abnormal observations via external memory construction (i.e., slot-based memory module).
For instance, R2Gen~\cite{chen2020generating} utilizes a memory matrix to memorize the projection between the visual patterns and language patterns, which is queried given the input image and fed to the decoder in the testing stage.
Similarly, a cross-modal memory network was proposed in~\cite{chen2021cross} to learn the latent features of abnormalities based on the visual features and on language features in the same latent space, aiming to effectively transform the cross modalities from visual to text. 

\section{An Overview of the Proposed Framework}
Given the radiology image with its extracted visual features denoted as $F^{(V)}$ as input, the objective is to generate a radiology report $R=\{y_1, y_2, ...\}$ where $y_i$ refers to the $i^{th}$ sentence in the report. We first introduce a fine-grained \textbf{AT}tributed \textbf{A}bnormality \textbf{G}raph (ATAG) to represent the relationships of abnormalities and their associated attributes, aiming to facilitate the generation of clinically accurate reports. 
We then propose a methodology to automatically construct a global 
ATAG $\mathcal{G}$ based on the given report corpus and the public radiology ontology. 
Given the ATAG structure, Graph Attention Network~\cite{velivckovic2017graph}) is adopted to learn the ATAG embedding $Z$ based on the input visual feature $F^{(V)}$.
By taking the ATAG embedding $Z$ as input, the decoder generates the final report $R$ by attending the proper attributed abnormality node embeddings in depicting different observations.

In the following sections, we first introduce the ATAG structure construction methodology (Section~\ref{sect:structure}) and ATAG embedding learning algorithms (Section~\ref{sect:embedding}). We then present how to integrate ATAG embedding for report generation with a hierarchical attention mechanism and a gate mechanism in Section~\ref{sect:integrate}.

\section{Abnormality Graph with Fine-grained Attributes}
\label{sect:structure}
In an X-ray medical report, items to be reported include the names of the abnormalities and their associated details such as the corresponding anatomical part, location, status, etc. To enable clinically accurate reports to be generated, 
we consider that it can only be possible if a more fine-grained abnormality graph representation can be constructed to represent the abnormality details. 
To represent such a graph, we first define $\mathcal{G}^{(A)}=(\mathcal{V}^{(A)}, \mathcal{E}^{(A)})$ where $\mathcal{V}^{(A)}$ represents a set of \textit{abnormality nodes} and $\mathcal{E}^{(A)}$ represents the set of edges connecting them. The abnormality nodes should be connected if they are inter-related. In addition, for each abnormality node $v^{(A)}_i \in \mathcal{V}^{(A)}$, it is paired with an attribute graph $\mathcal{G}^{(B_i)}=(\mathcal{V}^{(B_i)}, \mathcal{E}^{(B_i)})$ where $\mathcal{V}^{(B_i)}$ represent a set of associated \textit{attribute nodes} and $\mathcal{E}^{(B_i)}$ represents the set of edges connecting the attribute nodes (indicating that they are inter-related).

The attributed abnormality graph can therefore be denoted as $(\mathcal{G}^{(A)},\{\mathcal{G}^{(B_i)}\})$ with each $\mathcal{G}^{(B_i)}$ corresponding to a distinct $v^{(A)}_i \in \mathcal{V}^{(A)}$.

\subsection{Extracting abnormalities from X-ray annotations}
For the first step, we identify the set of abnormalities and their associated attributes to be included. Instead of manually identifying them as adopted in most of the existing methods, we propose to make use of the annotations as provided in the dataset. For instance, each image annotation in the IU XRay dataset typically contains terms about the abnormality and the associated descriptors (attributes). 
We adopt RadLex~(\texttt{http://radlex.org/})\footnote{
The RadLex Ontology is also employed in the chest X-ray report annotation guidance in IU XRay dataset~(\url{https://academic.oup.com/jamia/article/28/9/1892/6307885}).
} which is an ontology of radiology lexicon to extract i) the abnormality term $a$ if found under RadLex's ``clinical finding'' category, and ii) the associated attributes $\{b_{0},b_{1},...\}$ if found under RadLex's different descriptor categories. 
E.g., ``atelectasis'' is a ``clinical finding'' and ``right'' is a ``location descriptor'' in RadLex. 
For the annotation without any clinical finding term, we use ``other, [anatomical-part descriptor]'' to denote the abnormality. 
We applied this methodology to the IU X-Ray dataset and maintained terms of which the occurrence frequency larger than the certain threshold number. The extraction results are reported in Table.~\ref{tab:abnatr}

\begin{table}[]
\centering
\caption{The statistics of abnormalities and attributes appeared in the datasets. ``\textbf{Freq.}'' stands for the frequency threshold of the extracted terms.``\textbf{Abn.}'' and ``\textbf{Atr.}'' stand for the number of extracted abnormalities and attributes.}
\label{tab:abnatr}
\begin{tabular}{c|c|c|c|c}
\hline
\textbf{Dataset} & \textbf{Freq.} & \textbf{Abn.} & \textbf{Atr.} & \textbf{Max. / Min. /  Avg. / Std. per Abn.} \\
\hline
\multirow{3}{*}{\shortstack{IU \\ XRay}}  & 10  & 41 & 106 & 47 / 1 / 11.5 / 11.0  \\
  & 20  & 28 & 79  & 40 / 1 / 13.9 / 10.6  \\
  & 30  & 23 & 64  & 34 / 1 / 14.0 / 9.7 \\
\hline \hline 
\multirow{3}{*}{\shortstack{MIMIC \\ CXR}} & 500  & 47 & 209 & 178 / 17 / 69.1 /  38.6 \\
  & 1000 & 35 & 165 & 142 / 19 / 69.4 /  30.9 \\
  & 2000 & 26 & 129 & 116 / 37 / 70.0 /  22.4 \\ \hline
\end{tabular}

\end{table}

\subsection{Extracting abnormalities from X-ray reports}
\begin{figure}[htbp]
\centering
\includegraphics[width=0.45\textwidth]{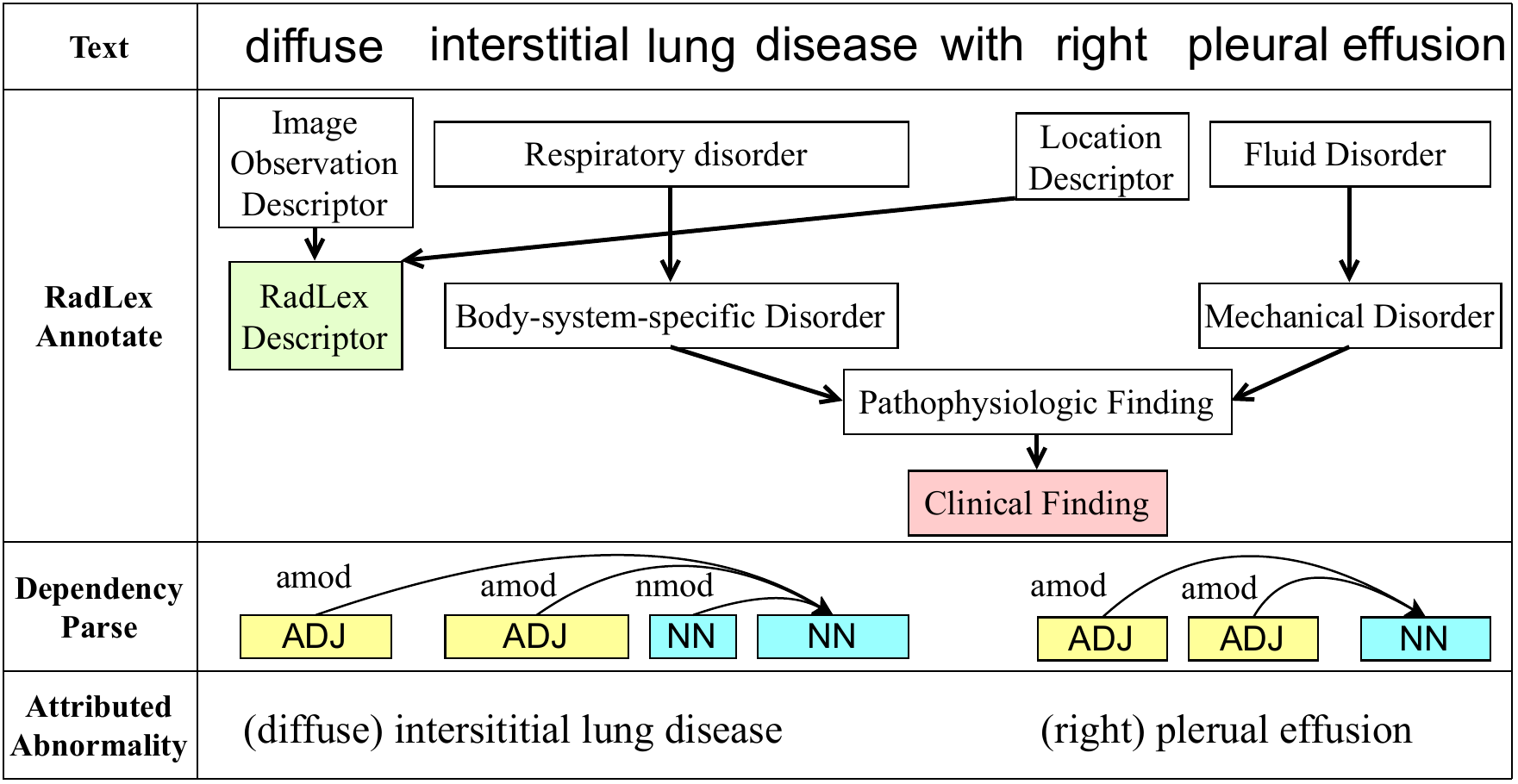}
\caption{Illustration of extracting attributed abnormality terms from free-text radiology reports using RadLex annotator and dependency parser.} 
\label{fig:radlex}
\end{figure}

We can also leverage some larger X-ray datasets (e.g., MIMIC CXR~\cite{johnson2019mimiccxr}) which have been made available recently. Very often detailed annotations are not provided since preparing ground-truth annotations is costly. Alternatively, we can make use of the X-ray reports provided in the dataset where information related to the abnormality and the associated descriptors can be extracted. An example is shown in Fig.~\ref{fig:radlex}. We can first filter out sentences of negative or inconclusive mentions using publicly available tools, i.e., clinical entity relationship parser RadGraph\footnote{RadGraph, a novel information extraction schema for radiology report information structuralization~ \url{https://physionet.org/content/radgraph/1.0.0/}}~\cite{jain2021radgraph} and radiology entity extraction RadLex-Annotator\footnote{\url{https://bioportal.bioontology.org/annotatorplus}}~\cite{martinez2017ncbo}.
The clinically related terms are first extracted from the reports using the open annotation API provided by RadLex.org~\cite{martinez2017ncbo}. 
The extracted terms under the ``Clinical Finding" category are used to form the abnormality nodes and ``RadLex Descriptor'' category are used to form the attribute nodes in ATAG.
Then, the dependency parser pre-trained on the MIMIC dataset (e.g., as in RadGraph~\cite{jain2021radgraph}) can be used to locate the attributes of the extracted abnormalities via the dependency relationships\footnote{We consider the parsed relationships ``located\_at'' and ``modify'' provided by RadGraph to determine the abnormality-to-attribute association.}. 
By applying this methodology to the MIMIC CXR dataset, different sizes of ATAGs are extracted as shown in Table.~\ref{tab:abnatr}.

\section{Learning Attributed Abnormality Graph Embedding}
\label{sect:embedding}
\begin{figure*}[h]
\centering
\includegraphics[width=0.95\textwidth]{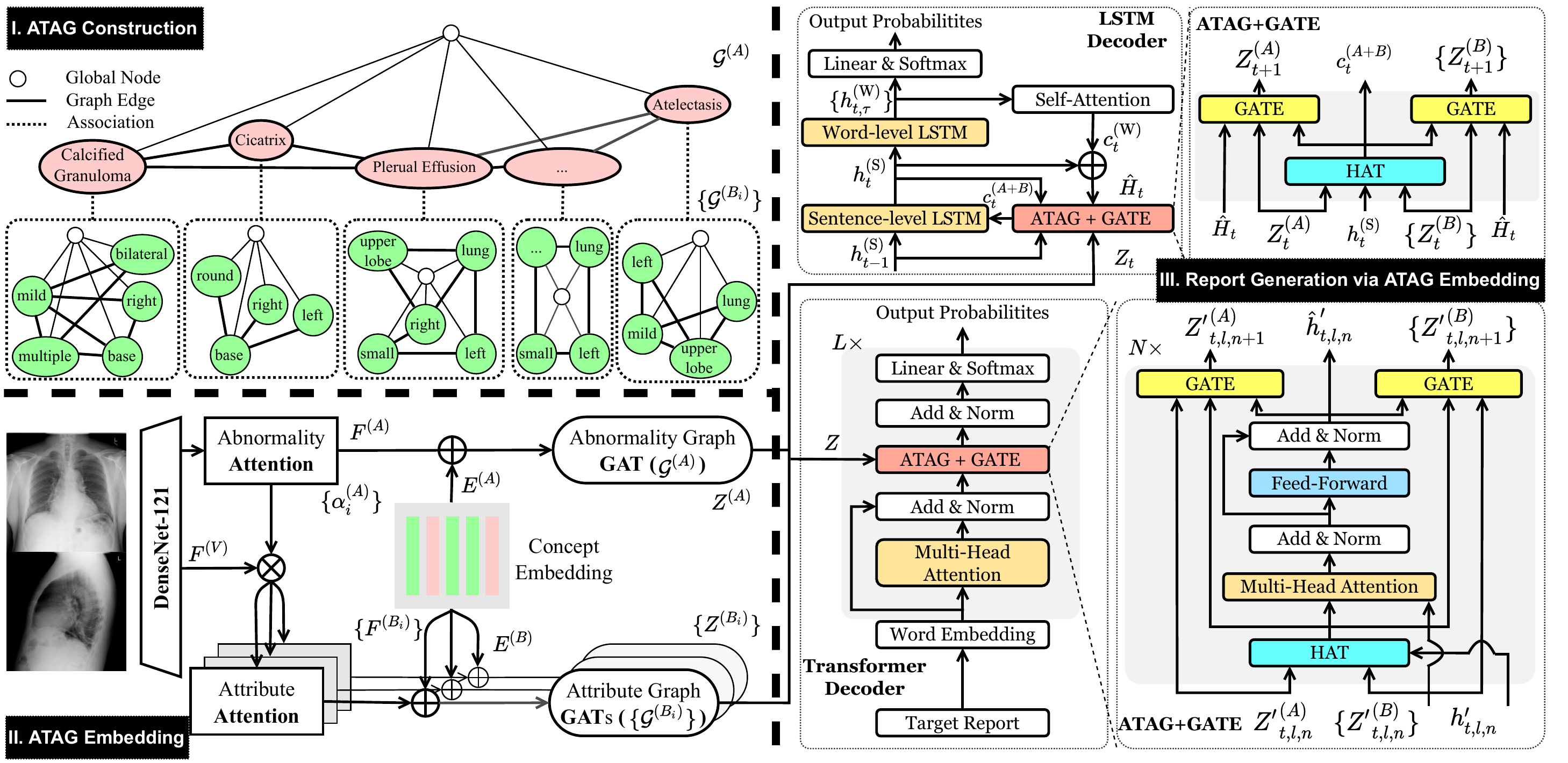}
\caption{The ATAG-based deep model architecture. An illustrated example of ATAG is presented in Part I, with the process of computing ATAG embedding shown in Part II and followed by the integration of ATAG and GATE with LSTM-based or Transformer-based decoder as depicted in Part III.} 
\label{fig:model}
\end{figure*}

We integrate the proposed ATAG into an encoder-decoder architecture similar to~\cite{zhang2020when}, as shown in Fig.~\ref{fig:model}. DenseNet~\cite{huang2017densely} is used to extract the visual feature for computing the ATAG embedding. 
Specific graph attentional layers (to be detailed) are introduced to aggregate the representations from heterogeneous nodes. The ATAG embeddings are learned with the multi-abnormality and multi-attribute classification as the learning objective. 

Given the visual features $F^{(V)} \in \mathbb{R}^{\mathcal{H} \mathcal{W} \times 2D}$ of size $\mathcal{H} \times \mathcal{W}$ extracted from the frontal and lateral chest x-ray images using the DenseNet-121 \cite{huang2017densely,zhang2020when}, we initialize the abnormality node features in ATAG using a spatial attention mechanism implemented by a convolutional layer $\mathrm{Conv}(\cdot)$. 
In particular, we set up $|A|$ channels with $1 \times 1$ filter size as $\mathrm{Conv}^{(A)} (F^{(V)}, |A|)$ such that each channel outputs $\alpha^{(A)}_i \in \mathbb{R}^{1 \times \mathcal{H} \mathcal{W}}$ as the attention weight to indicate the particular image region to be attended by node $v^{(A)}_i$. 
Then, the attending visual features $F^{(A)} \in \mathbb{R}^{|A|\times 2D}$ for $\mathcal{G}^{(A)}$ is computed by concatenating attention-weighted visual features
\begin{equation}
F^{(A)} = \bigoplus^{|A|}_{i=1}{\alpha^{(A)}_i F^{(V)}},
\end{equation}
where $\bigoplus$ denotes the concatenation operation. The visual feature for the global node $F^{(A)}_0$ is computed by the global average pooling of the visual features of all the other nodes. 

In addition, for all the abnormalities, we also define a set of intrinsic {\it concept embeddings} $E^{(A)}\in \mathbb{R}^{|A|\times D}$ to encode the a-priori information for the abnormalities and attributes. 
The {\it abnormality node embeddings}, denoted as $Z^{(A)} \in \mathbb{R}^{|A|\times D}$, are then computed based on the attending visual features $F^{(A)}$ and concept embeddings $E^{(A)}$ using the graph attentional layer~\cite{velivckovic2017graph} $\mathrm{GAT}(\cdot)$ on $\mathcal{G}^{(A)}$, given as
\begin{equation}
Z^{(A)} = \mathrm{GAT}((F^{(A)}\oplus E^{(A)})W_{Z^{(A)}}, \mathcal{G}^{(A)}),
\end{equation}
where $W_{Z^{(A)}} \in \mathbb{R}^{3D\times D}$ is a linear projection applied to reduce the dimension of the concatenated vector of the visual feature and concept embedding back to $D$.

For each attribute graph $\mathcal{G}^{(B_i)}$ associated with the abnormality node $v^{(A)}_i$, the visual features $F^{(V)}$ are first weighted by the channel output $\alpha^{(A)}_i$ and then the attribute attention is computed using another convolutional layer $\mathrm{Conv}^{(B_i)}$, where

\begin{equation}
\alpha^{(B_i)} = \mathrm{Conv}^{(B_i)} (\alpha^{(A)}_i F^{(V)}, |B_i|).
\end{equation}
The idea is to put focus on the region where the abnormality node is attending for computing the corresponding attribute embedding.
\noindent
Similar to modeling abnormality, we also define for all the possible attributes the corresponding set of intrinsic {\it concept embeddings} $E^{(B)}\in \mathbb{R}^{|B|\times D}$ where $B=\bigcup^{|A|}_{i=1}{B_{i}}$. The embedding of the attribute graph $Z^{(B_i)}\in \mathbb{R}^{|B_i|\times D}$ is then computed based on the concatenated attending visual features $F^{(B_i)} = \bigoplus^{|B_i|}_{k=1}{\alpha^{(B_i)}_k F^{(V)}}$ and the attribute concept embeddings $E^{(B_i)}\subseteq E^{(B)}$ with another linear projection $W_{Z^{(B)}}\in \mathbb{R}^{3D\times D}$ applied as
\begin{equation}
Z^{(B_i)} = \mathrm{GAT}(W_{Z^{(B)}}(F^{(B_i)}\oplus E^{(B)}), \mathcal{G}^{(B_i)}).
\end{equation}

For abnormality classification, $Z^{(A)}$ and $\{Z^{(B_i)}\}$ are fed to
a fully-connected layer with sigmoid functions employed. 
We learn the ATAG embedding end-to-end by optimizing the sum of the binary cross-entropy losses weighted by $w_i=|\mathcal{R}\setminus \mathcal{R}^{(a_i)}|/|\mathcal{R}^{(a_i)}|$ of each $a_i$, where $\mathcal{R}$ is the report set and  $\mathcal{R}^{(a_i)} \subseteq \mathcal{R}$ is the set of reports with $a_i$ mentioned in their annotations. Given the abnormality and attribute ground-truth labels, the total classification loss is defined as: 
\begin{equation}
\mathcal{L}_{\mathrm{CLS}} = \beta^{(A)} \mathcal{L}^{(A)}_{\mathrm{CLS}} + (1-\beta^{(A)}) \sum{
\beta^{(a_i)} * \mathcal{L}^{(B_i)}_{\mathrm{CLS}}},
\end{equation} 
where $\beta^{(A)}$ is a trade-off parameter between the abnormality and attribute loss functions, and $\beta^{(a_i)}= w_i/\sqrt{\sum^{|A|}_{k=1}{w^2_k}}$. 
In the training process, the global node $v^{(A)}_0$ is used to predict the existence of ``no finding / normal" label, and $v^{(B_i)}_0$ is used to predict the existence of any attribute labels associated with $a_i$.

\section{Report Generation with ATAG Embedding}
\label{sect:integrate}

After obtaining the abnormality graph embedding $Z^{(A)}$ and attribute graph embedding $\{Z^{(B_i)}\}$, a context vector $c^{(A+B)}$ is derived to guide the generation of the report $R$. 
With the objective to adaptively align specific information captured in $Z^{(A)}$ and $\{Z^{(B_i)}\}$ while different sentences in the report are being generated, we make reference to the decoder's hidden state (denoted as $h_t$) and propose a \textit{hierarchical attention mechanism} $\mathrm{HAT(\cdot)}$ for computing the attributed abnormality context vector, and a \textit{gating mechanism} $\mathrm{GATE(\cdot)}$ for adapting the abnormality graph embedding $Z^{(A)}$ and attribute graph embedding $\{Z^{(B_i)}\}$.
We will also show how the two mechanisms 
can be applied to LSTM- and Transformer-based decoders.

\subsection{Hierarchical Attention on ATAG Embeddings}


Given the decoder's current hidden state $h_t$ as the query, we can compute the attention to the attribute embeddings and then aggregate them. We then further compute the attention with reference to different abnormalities to implement the hierarchical attention.

We denote the overall ATAG \textit{context vector} at time step $t$ as $c^{(A+B)}_t$ which is computed by aggregating $Z^{(A)}$ and $\{Z^{(B_i)}\}$ using a hierarchical attention mechanism $\mathrm{HAT}(\cdot)$, given as 
\begin{equation}
c^{(A+B)}_t=\mathrm{HAT}(h_t, Z^{(A)},\{Z^{(B_i)}\}).
\end{equation}

\noindent \textbf{Aggregating attribute graph embedding with attention}. 
We first compute the aggregated context vector $c^{(B)}_{t}\in \mathbb{R}^{|A|\times D}$ of the attribute graphs as:
\begin{equation}
\begin{aligned}
\zeta^{(B_i)}_{t} &= \mathrm{Attention}^{(B)}(h_{t-1}, Z^{(B_i)}_{t}) \\
c^{(B)}_{t} &= \bigoplus^{|A|}_{i=1}{
\sum
\zeta^{(B_i)}_{t} 
Z^{(B_i)}_{t}
},
\end{aligned}
\end{equation}
where
$
\zeta^{(B_0)}_{t} = \mathrm{Attention}^{(B)}(h_{t-1},\bigoplus^{|A|}_{i=1}{Z^{(B_i)}_{t}})$ in which $\zeta^{(B_i)}_{t} \in \mathbb{R}^{|B_i|}$ and $\zeta^{(B_0)}_{t} \in \mathbb{R}^{\sum^{|A|}_{i=1}{|B_i|}}$, respectively, and 
$
\mathrm{Attention}(x,y) = 
\mathrm{softmax}(
 \mathrm{tanh}(
 xW_1+yW_2
 )W_3
 )
$
with $W_1, W_2 \in \mathbb{R}^{D\times D}$ and $W_3 \in \mathbb{R}^{D\times 1}$ being the learnable parameters.  
This aggregated attribute graph embedding aims to maintain the detailed information of clinical attributes, e.g., the positions ``\textit{left}'' and ``\textit{central}'' together with anatomical part ``\textit{subclavian}'' of medical device ``\textit{catheter tip}''.

\noindent \textbf{Attributed abnormality context vector}.
The attributed abnormality context vector $c^{(A+B)}_{t} \in \mathbb{R}^{2D}$ is then computed by combining the aggregated attribute context $c^{(B)}_{t}$ and abnormality embedding $Z^{(A)}_{t}$ as:
\begin{equation}
\begin{aligned}
\zeta^{(A)}_{t} &= \mathrm{Attention}^{(A)}(h_{t-1}, (Z^{(A)}_{t}\oplus c^{(B)}_{t})); \\
c^{(A+B)}_{t} &=\sum \zeta^{(A)}_{t} (Z^{(A)}_{t}\oplus c^{(B)}_{t}),
\end{aligned}
\end{equation}
where $\zeta^{(A)}_{t} \in \mathbb{R}^{|A|}$. 
The corresponding attention values are expected to indicate the attending abnormalities and attributes nodes for the generation of the next sentence or token. 
Note that the embedding of the global abnormality node is computed by global average pooling. 

\subsection{Gating Mechanism for Adaptive Graph Embeddings}
The knowledge graph embedding is expected to facilitate the report generation using the decoder, and it is mostly assumed to be unchanged while generating different abnormality observations~\cite{li2019knowledge,zhang2020when,liu2021exploring, liu2021auto}. 
As sentences in a radiology report are correlated, the sentences generated so far should affect the next sentence to be generated and the embedded knowledge required for the remaining decoding may also evolve accordingly. 
To facilitate 
this dynamic decoding process, we propose a gating mechanism $\mathrm{GATE}(\cdot)$ to allow the ATAG embedding $Z^{({A})}_{t}$ and $\{Z^{({B_i})}_{t}\}$ to evolve over time 
during the report generation. Given the current hidden state of the decoder $h_t$, the graph embedding $Z_{t}$ and the context vector $c_{t}$ as inputs, the gating mechanism is denoted as:
\begin{equation}
Z_{t+1} = \mathrm{GATE}(h_{t}, Z_{t}, c_{t}).
\end{equation}

To prevent $Z_{t}$ from gradient exploding due to the long sequence generation, we first compute an incremental graph embedding $\hat{Z}_{t}$
using a two-layer full connection neural network with residual connection by taking $Z_{t}$ and $c_{t}$ as the input, given as
\begin{equation}
\hat{Z}_{t} = 
\mathrm{FFN}(
Z_{t} 
+ c_{t}
)
+ Z_{t} 
+ c_{t}.
\end{equation} 
with

$
\mathrm{FFN}(X)=
\mathrm{GELU}(
X
W^{(\mathrm{FF})}_{1} + b^{(\mathrm{FF})}_{1})
W^{(\mathrm{FF})}_{2} + b^{(\mathrm{FF})}_{2}
$
where $\mathrm{GELU}(\cdot)$ is the activation function of Gaussian error linear units, $W^{(\mathrm{FF})}_{1} and W^{(\mathrm{FF})}_{2} \in \mathbb{R}^{D\times D}$ are the learnable weight matrices, $b^{(\mathrm{FF})}_{1}$ and $b^{(\mathrm{FF})}_{2}$ are the bias terms. 
$\hat{Z}_{t}$ is expected to differentiate the attended and non-attended embeddings in the $Z_t$ with reference to $c_t$.

In addition, to allow the knowledge embedding to be refined to ease the decoder for the subsequent decoding, a forget gate and an input gate are employed to determine the present and absent graph embeddings with reference to the hidden state, given as 
\begin{equation}
\begin{aligned}
O^{(\mathrm{F})}_{t} &= H_{t} W^{(\mathrm{F})}_{1} + \mathrm{tanh}(Z_{t})  W^{(\mathrm{F})}_{2}; \\
O^{(\mathrm{I})}_{t} &= H_{t} W^{(\mathrm{I})}_{1} + \mathrm{tanh}(Z_{t})  W^{(\mathrm{I})}_{2},
\end{aligned}
\end{equation}
where $H_{t}=\bigoplus^{|Z_{t}|}{h_{t}}$, and $W^{(\mathrm{I})}_{1}, W^{(\mathrm{I})}_{2}, W^{(\mathrm{F})}_{1},W^{(\mathrm{F})}_{2} \in \mathbb{R}^{D\times D}$ are the learnable parameters. 
Then, $Z_{t+1}$ is computed according to the forget context $O^{(\mathrm{F})}_{t}$ and input context $O^{(\mathrm{I})}_{t}$ as:
\begin{equation}
Z_{t+1} = 
\mathrm{Sigmoid}({O^{(\mathrm{I})}_{t})}\cdot Z_{t}
+ 
\mathrm{Sigmoid}({O^{(\mathrm{F})}_{t}})\cdot \mathrm{tanh}(\hat{Z}_{t}).
\end{equation}
The resulting $Z_{t+1}$ is expected to have some of the abnormality and attribute node embeddings already mentioned in the generated contents ``forgotten''. Such a refined graph embedding could progressively confine the number of the abnormality and attribute node embeddings to be attended by the decoder. 

\subsection{Decoding with Hierarchical LSTM}
Two-level long short-term memory (LSTM) network~\cite{graves2012long} is commonly used as the decoder for the report generation, where 
a top-level LSTM is used to predict the abnormalities (topics) for each sentence, and a bottom-level LSTM is to generate the description for the particular abnormalities. 
Given the ATAG context vector $c^{(A+B)}_t \in \mathbb{R}^{D}$, the top-level $\mathrm{LSTM}^{(\mathrm{S})}$ generates the sentence topic at step $t$ as:
\begin{equation}
\begin{aligned}
c^{(A+B)}_t &= \mathrm{HAT}(h^{(\mathrm{S})}_{t-1}, Z^{(A)},\{Z^{(B_i)}\}); \\
h^{(\mathrm{S})}_t &= \mathrm{LSTM}^{(\mathrm{S})} ( c^{(A+B)}_t, h^{(\mathrm{S})}_{t-1} )
\end{aligned}
\end{equation}
where the hidden state $h^{(\mathrm{S})}_{0}$ is initialized by concatenating the global average of abnormality and attribute node embeddings. 
The bottom level $\mathrm{LSTM}^{(\mathrm{W})}$ computes the hidden state $h^{(\mathrm{W})}_{t, \tau}$ for each word $w_{t, \tau}$ and generates the word sequence for the sentence as:
\begin{equation}
h^{(\mathrm{W})}_{t, \tau} = \mathrm{LSTM}^{(\mathrm{W})} (h^{(\mathrm{S})}_t \oplus c ^{(A+B)}_t, h^{(\mathrm{W})}_{t, \tau-1}).
\end{equation}
where each word is predicted by $y_{t,\tau}\sim p_{t,\tau} = \mathrm{softmax}(h^{(\mathrm{W})}_{t, \tau}W_y+b_y)$ in the generated report where $W_y$ and $b_y$ are learnable parameters.

\noindent \textbf{Integration with gating mechanism.}
The bottom-level $\mathrm{LSTM}^{(\mathrm{W})}$ generates the sequence of words guided by the topic information in $h^{(\mathrm{S})}_t$ of the top-level $\mathrm{LSTM}^{(\mathrm{S})}$ and the context vector $c^{(A+B)}_t$ attended by $h^{(\mathrm{S})}_t$. 
However, the actual generated content only covers parts of the topic, i.e., $\{h^{(\mathrm{W})}_{t, \tau}\} \subset (h^{(\mathrm{S})}_t\oplus c^{(A+B)}_t)$. 
The inconsistency between the two-level LSTMs limits the decoder to generate the completed descriptions of detected abnormalities and attributes. 
In addition, each sentence is generated by attending proper attributed abnormality embedding from a large volume of embeddings\footnote{For example, ATAG (41+106) for IU XRay has 41 abnormality nodes and 472 attribute nodes in total, where each node is associated with a embedding vector.}. 
The effectiveness of attention module is expected to be improved if the values of relevant node embeddings are boosted.

To supplement the attended but not generated context $(h^{(\mathrm{S})}_t\oplus c^{(A+B)}_t)/\{h^{(\mathrm{W})}_{t, \tau}\}$ in generating the next sentences and enhance the attention effectiveness, the graph embedding is expected to increase the chances for the corresponding graph embedding for the decoder.
The structure is illustrated by Fig.~\ref{fig:model}. 
Thus, after decoding each sentence as Eq.(13), the generation module is followed by operating the gate mechanism $\mathrm{GATE}(\cdot)$ to update the graph embedding as,  
\begin{equation}
\begin{aligned}
Z^{(A)}_{t+1} = \mathrm{GATE}^{(A)}(H_{t}, Z^{(A)}_{t}, c^{(A)}_{t}); \\
Z^{(B)}_{t+1} = \mathrm{GATE}^{(B)}(H_{t}, Z^{(B)}_{t}, c^{(B)}_{t}).
\end{aligned}
\end{equation}
where $c^{(A)}_{t} = \bigoplus \zeta^{(A)}_{t} Z^{(A)}_{t}$ is the individual context vector of $Z^{(A)}_t$.
The aggregated hidden state $H_{t}\in \mathbb{R}^{D}$ is computed by the outputs of both sentence- and word-level LSTMs $h^{(\mathrm{S})}_t$ and $\{h^{(\mathrm{W})}_{t, \tau}\}$ as, 
\begin{equation}
\begin{aligned}
H_{t} &= (h^{(\mathrm{S})}_t \oplus c^{(\mathrm{W})}_{t})W^{(\mathrm{H})} ;\\
c^{(\mathrm{W})}_{t} &= 
\bigoplus_{\tau=1}
\zeta^{(\mathrm{W})}_{t, \tau}
h^{(\mathrm{W})}_{t, \tau} 
; \\
\zeta^{(\mathrm{W})}_{t, \tau} &= 
\mathrm{Attention}^{(\mathrm{W})}
( h^{(\mathrm{W})}_{t, \tau},
\{h^{(\mathrm{W})}_{t, \tau}\})
\end{aligned}
\end{equation}
where $W^{(\mathrm{H})}\in \mathbb{R}^{2D\times D}$ is the linear projection. $h^{(\mathrm{S})}_t$ from the top level $\mathrm{LSTM}^{(\mathrm{S})}$ aims to maintain the information of attended attributed abnormality embedding that is expected to generate at $t$-th sentence
Meanwhile, $c^{(\mathrm{W})}_{t}$ is expected to retain the detailed descriptions of particular attributes in that sentences.  
The updated $Z^{(A)}_{t+1}$ and $\{Z^{(B)}_{t+1}\}$ are then expected to enhance $\mathrm{LSTM}^{(\mathrm{S})}$ to look up the proper context embedding in the following generation.
The chances of accurately describing abnormalities and attributes in the report are thus increased.

\subsection{Decoding with Transformer}
To generate the proper report by attending the corresponding graph emebeddings, we also integrate the proposed ATAG with the effective Transformer-based decoder. The Transformer-based decoder is constructed by multi-head attention (MHA)~\cite{vaswani2017attention} which is 
composed by multiple parallel the scaled dot-product attention modules.
Given the mixed embedding of each word $x_{t, 0} = w_t + e_t$ by word embedding $w_t$ and positional embedding $e_t$, the decoder aims to generate each word $\hat{y}_t$ with $L$-layer MHA. 
For $l$-th layer, the hidden state $h_{t, l}$ is computed by $x_{t, l-1}$, 
\begin{equation}
h_{t, l} =\mathrm{MHA}(x_{t, l-1},x_{1:t, l-1}),
\end{equation}
Then, $\hat{h}_{t,l}$ is decoded by $Z^{(A)}$ and $\{Z^{(B)}\}$ with $\mathrm{HAT}(\cdot)$ and $\mathrm{MHA}(\cdot)$ as, 
\begin{equation}
\begin{aligned}
c_{t, l}^{(A+B)} &= \mathrm{HAT}(h_{t,l}, Z^{(A)}, \{Z^{(B_i)}\}) \\
\hat{h}_{t,l} &= \mathrm{FFN}(\mathrm{MHA}(h_{t,l}, c_{t, l}^{(A+B)})), 
\end{aligned}
\end{equation}
where $\mathrm{MHA}(\cdot)$ and $\mathrm{FFN}(\cdot)$ are followed by dropout~\cite{srivastava2014dropout}, skip connection~\cite{he2016deep} and layer normalization~\cite{ba2016layer} to alleviate the data bias. 
We use the last layer output $\hat{h}_{t,L}$ to predict each word $y_t$.

\noindent \textbf{Integration with gating mechanism.} 
The hidden state $\hat{h}_{t}$ is generated by attending the preceding token sequence $\{x_t\}$ which covers multiple observations in the multiple sentences generated. 
Thus, evolving graph embedding $Z_t$ by token-level $h_{t}$
would cause $Z_t$ to ``forget'' both attended node embeddings in the preceding sentences $h_{1:t}$
and non-attended node embeddings that would be attended in the current sentence $h_{t+1:T}$. 

To evolve the graph embedding by the actually attended attributed abnormalities in the generated content, the corresponding fine-grained $\hat{h}_{t,l,n}^{\prime} \subset \hat{h}_{t,l}$ is first to be computed by $N$-times recursion inside each Transformer layer.
As illustrated in Fig.~\ref{fig:model}, for $n$-th recursion in $l$-th layer, the context vector of attributed abnormality embedding ${c^{\prime}}_{t,l,n}^{(A+B)}$ is computed as,
\begin{equation}
\begin{aligned}
{c^{\prime}}_{t,l,n}^{(A+B)} &= \mathrm{HAT}(h_{t,l,n}^{\prime}, {Z^{\prime}}^{(A)}_{t,l,n}, \{{Z^{\prime}}^{(B)}_{t,l,n}\}); \\
\hat{h}_{t,l,n}^{\prime} &= \mathrm{FFN}(\mathrm{MHA}(h_{t,l,n}^{\prime}, c_{t,l, n}^{(A+B)})), 
\end{aligned}
\end{equation}
where $h_{t,l,0}^{\prime}=h_{t,l}$, ${Z^{\prime}}^{(A)}_{t,l,0}=Z^{(A)}$ and ${Z^{\prime}}^{(B)}_{t,l,0}=Z^{(B)}$. 
For $(n+1)$-th recursion, $h_{t,l,n+1}^{\prime}$ is initialized by $\hat{h}_{t,l,n}^{\prime}$, and the last recursion $\hat{h}_{t,l,N}^{\prime}$ is taken as the output $\hat{h}_{t,l}$ for $l$-th layer. 
Accordingly, the graph embedding is accordingly evolved by $\mathrm{GATE}(\cdot)$ as, 

\begin{equation}
\begin{aligned}
{Z^{\prime}}^{(A)}_{t,l,n+1} &= \mathrm{GATE}(\hat{h}_{t,l, n}, c^{(A)}_{t,l,n}, {Z^{\prime}}^{(A)}_{t,l,n}); \\
{Z^{\prime}}^{(B)}_{t,l,n+1} &= \mathrm{GATE}(\hat{h}_{t,l, n}, c^{(B)}_{t,l,n}, {Z^{\prime}}^{(B)}_{t,l,n}).
\end{aligned}
\end{equation}
where ${Z^{\prime}}^{(A)}_{t,L,N}$ and $\{{Z^{\prime}}^{(B)}_{t,L,N}\}$ are taken as the $Z^{(A)}_{t+1}$ and $\{Z^{(B)}_{t+1}\}$ for the following generation. 
In this way, the gating mechanism will be performed $L\times N$ times to refine graph embedding in a fine-grained manner. 
It is expected to enhance the Transformer decoder to generate the accurate descriptions of abnormalities and associated attributes.

\section{Clinical Accuracy Evaluation}
To measure the clinical accuracy of the generated radiology reports in a more fine-drained manner,
we propose a new metric \textbf{Rad}iology \textbf{R}eport \textbf{Q}uality \textbf{I}ndex (RadRQI) to evaluate the accuracy of radiology-related abnormalities with the clinical attributes.

Given a radiology report, the radiology-related terms of abnormality and their attributes are first extracted using the RadLex ontology~\cite{martinez2017ncbo}\footnote{An open annotation API is provided by RadLex.org at \url{https://bioportal.bioontology.org/annotatorplus}} 
The keywords if found under RadLex's category 
i) ``\textit{Clinical Finding}'' will be taken as abnormalities, 
and ii) ``\textit{Clinical Descriptor}'' will be taken as clinical attributes. An example is shown in Fig.~\ref{fig:example}. 
Next, the negation and association of each abnormality or attribute is determined 
using the RadGraph, an entity and relation parser
trained by CheXpert~\cite{irvin2019chexpert} and MIMIC CXR~\cite{johnson2019mimiccxr} radiology report datasets. 
The clinical-related terms will be taken as ``Positive'' if labeled as ``Definitely Present'', and ii) ``Negative'' if labeled as ``Definitely Absent''. 
The association relationship from attribute to abnormality is determined if the ``Modify'' or ``Located At'' relation is detected. 
As a result, the tuples of the form ``(Abnormality, Negation, [Attribute1, Attribute2, ...])'' are extracted for the subsequent RadRQI score calculation. 
Similar to previous work~\cite{demner2016preparing,chen2020generating,miura2021improving}, the probable (but not definite) existence of clinical findings are not considered. 

By extracting the abnormalities together with their attributes from both the generated and ground truth reports, the precision, recall and F-1 measure scores are computed for each abnormality category. 
For each ``positive'' mentioned abnormality, similar to~\cite{zhang2020when}, the number of True Positives (TP) considers also the correct hits of the corresponding attributes, 
\begin{equation}
\mathrm{TP} = (1-\alpha^{(B)}) \mathrm{TP}^{(A)} + \alpha^{(B)} \mathrm{TP}^{(B)}
\end{equation}
where $\alpha^{(B)}$ is the weight of the attribute term accuracy to determine its contribution in the overall TP calculation.
The proposed RadRQI-F1 score aims to reflect the correctness of mentioned abnormality with associated attributes. 
In addition, the number of abnormality categories with non-zero F1 score, denoted as RadRQI-Hits, is also reported to show the coverage of distinct abnormality categories in the generated reports. 
To avoid rare abnormalities, we compute RadRQI by considering only top-$K$ abnormalities in terms of their frequencies in the datasets.

\begin{figure}[!t]
\centering
\includegraphics[width=0.30\textwidth]{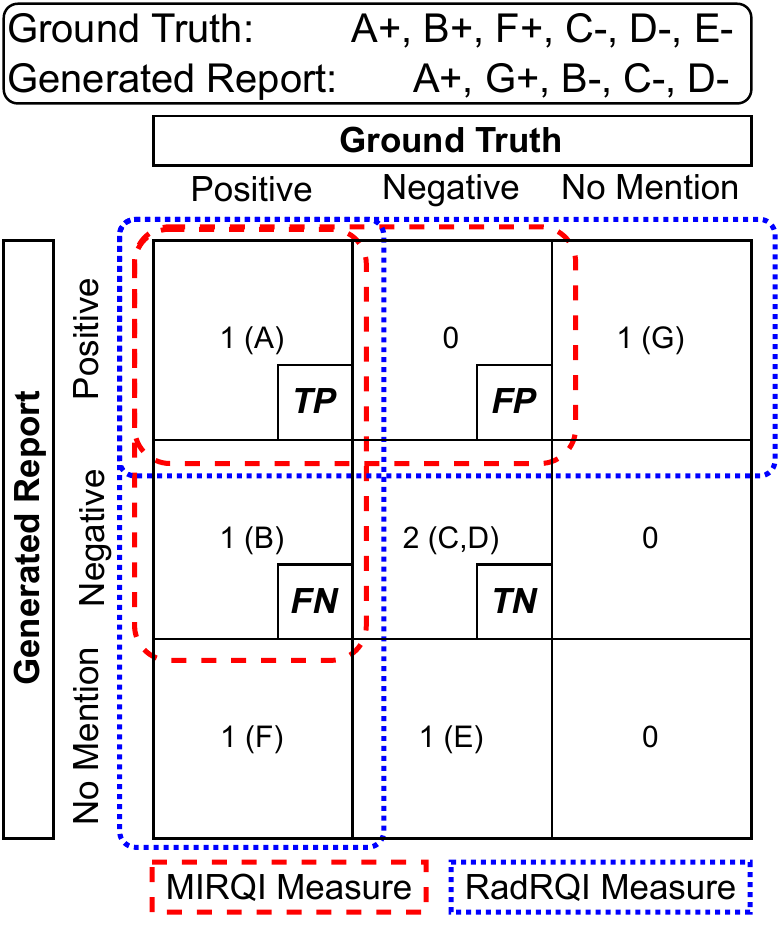}
\caption{Illustration of MIRQI and the proposed RadRQI for calculating the precision and recall of a generated report with two positive mentions and three negative mentions of abnormalities.} 
\label{fig:radrqi}
\end{figure}

Noted that
Medical Image Report Quality Index (MIRQI), proposed in ~\cite{zhang2020when}, also measures the correctness of attributes of the mentioned abnormalities in the generated reports. 
However, MIRQI evaluates a small set of abnormalities which covers 12 disease categories (labeled by CheXpert labeling toolkit) and takes irrelevant words (e.g., stop words like ``is'' and ``no'') as the corresponding attributes of some abnormalities.
Meanwhile, MIRQI does not consider terms which are found in the ground truth but not mentioned in the generated report, nor those found in the generated reports but not mentioned in the ground truth. By ignoring those \textit{not-mentioned} terms in the evaluation, it will favor methods which keep generating only a few correct abnormalities but missing many others, thus resulting in misleading evaluation results.
As illustrated in Fig.~\ref{fig:radrqi}, by ignoring the not-mentioned terms, the precision and recall calculated by MIRQI is $\mathrm{TP}/(\mathrm{TP}+\mathrm{FP})=1/(1+0)=1$ and $\mathrm{TP}/(\mathrm{TP}+\mathrm{FN})=1/(1+1)=0.5$, respectively. 
While for RadRQI, counting also the not-mentioned terms, the precision and recall become $1 (\mathrm{A})/(1 (\mathrm{A})+1 (\mathrm{G}))=0.5$ and $1 (\mathrm{A})/(1 (\mathrm{A})+1(\mathrm{B})+1 (\mathrm{F}))=0.33$, respectively. 
Comparing with the evaluation results of MIRQI which gives a high score to the partially correct generated report, RadRQI gives more reliable evaluation of the medical term accuracy in the generated report. 

\section{Experiments}

\subsection{Datasets and Evaluation Metrics}

We use two publicly available datasets IU X-Ray~\cite{demner2016preparing} and MIMIC CXR~\cite{johnson2019mimiccxr} for performance evaluation. 
The statistics of the datasets are shown in Table.~\ref{tab:datasets}.
For the IU X-Ray dataset, 
similar to~\cite{li2019knowledge,zhang2020when}, we extract only the reports with both frontal and lateral view images, complete finding/impression sections and annotations available, resulting 2,848 cases and 5,696 images. 
We tokenize all the words in the reports and filter out tokens with frequency less than three, resulting in 1,028 unique tokens.
We partition the data into training/validation/test set by 7:1:2 for five-fold cross validation. 
For the MIMIC CXR dataset, we apply an open source tool \footnote{\url{https://github.com/abachaa/MEDIQA2021/tree/main/Task3}} to extract findings/impression sections as the target report and filter out tokens with frequency less than 10, resulting 4,936 distinct tokens and following the original split set with training/validation/test size as 222,705 / 1,807 / 3,269. 
We report the average performance scores of three different runs.
\begin{table}[!t]
\caption{The statistics of the datasets used in our experiments.}
\label{tab:datasets}
\centering
\begin{tabular}{l|c|c}
\hline 
\textbf{Dataset} & \textbf{IU XRay~\cite{demner2016preparing}} & \textbf{MIMIC CXR~\cite{johnson2019mimiccxr}} \\
\hline 
Image \#  & 7,470 & 473,057 \\
Report \# & 3,996 & 206,563 \\
Case \#  & 3,996 & 63,478  \\
Avg. Len. & 38.3  & 53.2 \\
Avg. Sentence \#  & 5.8  & 5.5  \\
Avg. Sentence Len. & 6.5  & 10.8  \\
\hline 
\end{tabular}

\end{table}

Regarding evaluation metrics, 
the AUC of Receiver Operating Characteristic (ROC) curve and Precision Recall (PR) curve are used for measuring the multi-label classification performance. The micro-average score is reported.
For report quality, we adopt the common natural language generation metrics like BLEU~\cite{papineni2002bleu}, ROUGE~\cite{lin2004rouge} and CIDEr~\cite{vedantam2015cider} which measure the similarity between the generated report and the ground truth. 
For clinical accuracy of the generated report, we adopt the clinical efficacy metrics (CE)~\cite{chen2020generating} and its modifications~\cite{miura2021improving,li2019knowledge} to evaluate the accuracy of a series of observation presence status comparing with the ground truth. We use the CheXpert labeling toolkit\footnote{\url{https://github.com/MIT-LCP/mimic-cxr/tree/master/txt/chexpert}} to label 12 different thoracic diseases together with ``medical device'' and ``normality''. The micro-average F1 scores and the average number of classes which have non-zero F1 scores are reported, denoted as \texttt{CE(Hits)}.
We also adopt the proposed RadRQI metric to evaluate the clinical accuracy of a large number of abnormalities and attributes. We focus on more common abnormalities and attributes, and thus measure the RadRQI scores for top-25 and top-50 for IU XRay and MIMIC CXR, respectively.
The $\alpha^{(B)}$ is set to 0.5 indicating the equal importance of abnormalities and their associated attributes. 

\subsection{Baselines for Performance Comparison}
We first evaluate the performance of ATAG for the multi-label classification. We compare variants of ATAG with a number of baselines where DenseNet~\cite{huang2017densely} is adopted for the visual feature extraction but with different encoders and different number of labels considered, denoted as
\texttt{DenseNet[+Encoder]([\# Labels]}). 
The encoders tested include fully-connected layer (DenseNet)~\cite{irvin2019chexpert}, knowledge graph (DenseNet+KG)~\cite{zhang2020when}, the abnormality graph in ATAG (DenseNet+AG), and the ATAG with both the abnormality graph and the set of associated attribute graphs (DenseNet+ATAG). 
Regarding the labels, ``(20)'' refers to the 20 labels used in~\cite{zhang2020when}. 
For IU X-Ray dataset, ``(41)'' corresponds to the 41 abnormalities in ATAG, and ``(41+106)'' to further 106 distinct attributes included. For MIMIC CXR dataset, ``(47)'' refers to the 47 abnormalities in ATAG, and ``(47+209)'' to further 209 distinct attributes included. \texttt{DenseNet+KG(20)} is equivalent to
\cite{zhang2020when}, and \texttt{DenseNet+ATAG(41+106)} and \texttt{DenseNet+ATAG(47+209)} refer to our proposed methods. 
All input images are resized to $512 \times 512$ before feeding into the DenseNet and no normalizing pre-processing is further adopted.

To evaluate the effectiveness of the ATAG-based approach for report generation, we integrate ATAG with both LSTM-based decoders and Transformer-based decoders. For the LSTM-based decoders, we tested five state-of-the-art LSTM-based report generation models as the baselines, including the classical {CNN-RNN} model \texttt{WordSAT}~\cite{Xu2015Show} with a one-level LSTM decoder, \texttt{AdaAttn} model~\cite{lu2017knowing} with an adaptive attention module and a one-level LSTM decoder, \texttt{SentSAT}~\cite{yuan2019automatic} with a two-level LSTM decoder, \texttt{CoAtt}~\cite{jing2018automatic} with additional label features in addition to \texttt{SentSAT}, and \texttt{SentSAT+KG}~\cite{zhang2020when} which utilizes a knowledge graph with 20 abnormalities. Accordingly, \texttt{SentSAT+ATAG} and \texttt{SentSAT+ATAG+GATE} refer to the basic two-level LSTM integrated with our proposed ATAG and GATE modules. 
For the Transformer-based decoders, we integrate ATAG with the basic Transformer (\texttt{Trans+ATAG}) only and also together with the proposed GATE module (\texttt{Trans+ATAG+GATE}). We compare the performance of the proposed ATAG-based Transformer models 
with the vanilla Transformer \texttt{Transformer}, the state-of-the-art image captioning model \texttt{M2 Trans}~\cite{cornia2020meshed} with memory-enhanced Transformer encoder\footnote{\url{https://github.com/aimagelab/meshed-memory-Transformer}} and two open source report generation models  
\texttt{R2Gen}\footnote{\url{https://github.com/cuhksz-nlp/R2Gen}}~\cite{chen2020generating} and \texttt{R2Gen-CMN}\footnote{\url{https://github.com/cuhksz-nlp/R2GenCMN}}~\cite{chen2021cross}. 

\subsection{Experiment Settings}
We adopt the DenseNet-121 pretrained on the CheXpert dataset\footnote{\url{https://github.com/gaetandi/cheXpert}} as the visual encoder.
We use the implementation in the deep graph Python library\footnote{\url{https://github.com/dmlc/dgl}} for the GAT used in our graph embedding learning.
For the report generation, the dimension of the hidden states in all LSTM decoders is 512.
For the Transformer-based decoders, the dimensions of hidden states, the number of heads, the number of layers and the number of looping are set to be 512, 8, 2 and 3 respectively.
Two-phrase training is adopted where the encoder is trained and then fixed during the training of the decoder~\cite{zhang2020when}.
For IU XRay, the encoder is trained with the learning rate 1e-6 for 150 epochs, followed by the decoder with the learning rate 1e-4 for 100 epochs. 
The mini-batch size for the training is 8. 
For MIMIC CXR, the encoder and decoder are trained for 32 epochs using the mini-batch size of 16 and the learning rates 1e-6 and 1e-4, respectively. 

\subsection{Performance on Multi-label Classification}
We report the ROC-AUC and PR-AUC scores of all the models for comparing the classification accuracy, as shown in Table~\ref{tab:cls}.
The models trained using the abnormalities and attributes we extracted give significantly better prediction results. 
Also, the use of ATAG gives the best ROC-AUC and PR-AUC scores on average, implying the effectiveness of incorporating the attributed abnormality graph for the classification. 
Comparing the models with and without ATAG, it is clear that the attributes introduced in ATAG do lead to accuracy improvement.

\begin{table}[!t]
\caption{Performance on multi-label classification (AUC) over all the categories being trained. ``Abn.'' and ``Atr.'' stand for abnormality classification and attribute classification, respectively.}
\label{tab:cls}
\centering
\begin{tabular}{c|l|c|c}
\hline 
\textbf{Dataset} & \textbf{Model} & \textbf{Abn.} & \textbf{Atr.} \\
\hline 
\multirow{14}{*}{\shortstack{IU \\ XRay}} & \multicolumn{3}{l}{\textbf{ROC-AUC (std.)}} \\ \cline{2-4}
 & DenseNet (20)~\cite{irvin2019chexpert} & \textbf{0.740}$\pm$0.019 & - \\
 & DenseNet + KG (20)~\cite{zhang2020when} & 0.728$\pm$0.002 & - \\ \cline{2-4}
 & DenseNet (41) & \textbf{0.890}$\pm$0.009 & - \\
 & DenseNet + AG (41) & 0.888$\pm$0.003 & - \\ \cline{2-4}
 & DenseNet (41+107) & 0.884$\pm$0.012 & 0.560$\pm$0.054 \\
 & DenseNet + ATAG (41+107) & \textbf{0.892}$\pm$0.006 & \textbf{0.686}$\pm$0.069 \\ \cline{2-4}
 & \multicolumn{3}{l}{\textbf{PR-AUC}} \\ \cline{2-4}
 & DenseNet (20)~\cite{irvin2019chexpert} & 0.092$\pm$0.024 & - \\
 & DenseNet + KG (20)~\cite{zhang2020when} & \textbf{0.595}$\pm$0.103 & - \\ \cline{2-4}
 & DenseNet (41) & 0.793$\pm$0.099 & - \\
 & DenseNet + AG (41) & \textbf{0.795}$\pm$0.102 & - \\ \cline{2-4}
 & DenseNet (41+107) & 0.801$\pm$0.109 & 0.530$\pm$0.104 \\
 & DenseNet + ATAG (41+107) & \textbf{0.810}$\pm$0.110 & \textbf{0.799}$\pm$0.132 \\
\hline \hline
\multirow{10}{*}{\shortstack{MIMIC \\ CXR}} &
\multicolumn{3}{l}{\textbf{ROC-AUC (std.)}} \\ \cline{2-4}
 & DenseNet (47) & 0.897$\pm$0.001 & - \\
 & DenseNet + AG (47) & \textbf{0.916}$\pm$0.005 & - \\ \cline{2-4}
 & DenseNet (47+209) & 0.894$\pm$0.003 & 0.565$\pm$0.031 \\
 & DenseNet + ATAG (47+209) & \textbf{0.907}$\pm$0.006 & \textbf{0.683}$\pm$0.058 \\
\cline{2-4}
& \multicolumn{3}{l}{\textbf{PR-AUC (std.)}} \\
\cline{2-4}
& DenseNet (47) & 0.510$\pm$0.088 & - \\
 & DenseNet + AG (47) & \textbf{0.519}$\pm$0.120 & - \\ \cline{2-4}
 & DenseNet (47+209) & 0.513$\pm$0.103 & 0.440$\pm$0.201 \\
 & DenseNet + ATAG (47+209) & \textbf{0.529}$\pm$0.090 & \textbf{0.509}$\pm$0.135 \\
\hline 
\end{tabular}
\end{table}

\subsection{Performance on Report Generation}

\begin{table*}[!t]
\caption{Performance comparison of report generation models evaluated by two clinical accuracy metrics and NLG metrics. ``Top-K'' are set to top-25 and top-50 in IU XRay and MIMIC CXR, respectively. The best scores are in bold face and the second best are underlined. ``B.'', ``R.'' and ``C.'' stand for BLEU, ROUGE and CIDEr scores. }
\label{gen:all}
\centering
\begin{tabular}{c|l|cccc|cccc||ccc}
\hline
\multirow{2}{*}{\textbf{Dataset}} & \multicolumn{1}{l}{\multirow{2}{*}{\textbf{Model}}} & \multicolumn{4}{|c}{\textbf{Clinical Efficacy}}  & \multicolumn{4}{|c||}{\textbf{RadRQI}} & \multicolumn{3}{|c}{\textbf{NLG}} \\
& & \textbf{(5)} & \textbf{(14-1)} & \textbf{(14)} & \textbf{Hits}  & \textbf{(5)} & \textbf{(14-1)} & \textbf{Top-K} & \textbf{Hits} & \textbf{B.} & \textbf{R.} & \textbf{C.} \\
\hline
\multirow{21}{*}{\shortstack{IU \\ XRay}} &  \multicolumn{9}{l}{\textbf{LSTM-based Model}} \\  \cline{2-13}
& WordSAT (20)~\cite{Xu2015Show}  & 0.085 & 0.074  & 0.175 & 4.8  & 0.019 & 0.018  & 0.024 & 3.6  & 0.262 & 0.369 & 0.317 \\
 & SentSAT  (20)~\cite{yuan2019automatic} & 0.087 & 0.083  & 0.171 & 5.6  & 0.012 & 0.013  & 0.030 & 5.8 & 0.261 & 0.363 & 0.344\\
 & CoAttn  (20)~\cite{jing2018automatic}  & 0.056 & 0.068  & 0.167 & 5.2  & 0.009 & 0.013  & 0.023 & 5.0 & {0.274} & 0.365 & 0.318\\
 & SentSAT+KG (20)~\cite{zhang2020when}  & 0.061 & 0.069  & 0.173 & 4.8  & 0.012 & 0.012  & 0.024 & 3.6 & \underline{0.275} & \textbf{0.374} & 0.351\\ \cline{2-13}
 & WordSAT  (41)~\cite{Xu2015Show} & 0.194 & 0.140  & 0.249 & 5.6  & \underline{0.074} & 0.065  & 0.060 & 6.4 & 0.267 & \underline{0.369} & \textbf{0.359}\\
 & AdaAttn  (41)~\cite{lu2017knowing} & \underline{0.203} & 0.147 & 0.258 & 6.6 & {0.070} & \underline{0.066}  & 0.068 & 7.6 & 0.269 & 0.367 & \underline{0.358}\\
 & SentSAT  (41)~\cite{yuan2019automatic} & \textbf{0.223} & \underline{0.164} & \textbf{0.268} & \underline{7.6} & 0.067 & 0.064  & 0.061 & 9.8 & 0.272 & 0.362 & 0.326 \\
 & CoAttn  (41)~\cite{jing2018automatic}  & 0.143 & 0.108  & 0.220 & 5.8 & 0.046 & 0.045 & 0.055 & 6.8 & 0.259 & 0.364 & 0.340 \\
 & SentSAT  (41+106)~\cite{yuan2019automatic}  & 0.157 & 0.123  & 0.229 & 5.6 & 0.052 & 0.048  & 0.056 & 6.4 & 0.261 & 0.357 & 0.307 \\ \cline{2-13}
 & SentSAT+AG (41) & 0.164 & 0.110 & 0.227 & 4.8 & \textbf{0.078} & 0.054 & 0.043 & 4.6 & \textbf{0.323} & \textbf{0.374} & 0.297 \\
 & SentSAT+ATAG (41+106) & 0.190 & 0.145 & 0.244 & 7.2 & 0.062 & 0.065  & \underline{0.069} & \underline{10.2} & 0.255 & 0.351 & 0.356 \\
 & SentSAT+ATAG+GATE (41+106) & \underline{0.216} & \textbf{0.178}  & \underline{0.263}  & \textbf{8.6}  & 0.066 & \textbf{0.068}  & \textbf{0.079} & \textbf{12.4} & 0.264 & 0.349 & 0.349 \\ \cline{2-13}
& \multicolumn{9}{l}{\textbf{Transformer-based Model}} \\  \cline{2-13}
& Transformer~\cite{vaswani2017attention}  & 0.124 & 0.112 & \textbf{0.310} & 5.0  & {0.052}  & 0.038 & \underline{0.072} & 9.0 & \underline{0.264} & 0.357 & \underline{0.587} \\
  & M2 Trans.~\cite{cornia2020meshed} & 0.130 & 0.111  & 0.205 & 8.0  & 0.029 & 0.030  & 0.040 & 8.0 & 0.255 & \underline{0.367} & 0.313 \\
  & R2Gen~\cite{chen2020generating} & 0.115 & 0.127  & 0.289 & 9.0  & 0.040 & \textbf{0.057}  & 0.071 & 10.0 & 0.251 & 0.342 & 0.461\\
  & R2Gen-CMN~\cite{chen2021cross} & 0.098 & 0.121 & \underline{0.290} & 8.0  & 0.034 & 0.037  & 0.056 & 10.0 & \textbf{0.294} & \textbf{0.370} & \textbf{0.681} \\ \cline{2-13}
  & Trans.+AG (41) & 0.207 & 0.179 & 0.277 & 9.6 & \textbf{0.063} & 0.054 & 0.069 & 6.4 & 0.256 & 0.357 & 0.304 \\
  & Trans.+ATAG (41+106) & \underline{0.184} & \underline{0.178} & 0.262 & \underline{10.0} & 0.050 & \underline{0.047} & \underline{0.072} & \textbf{13.8} & 0.246 & 0.334 & 0.334\\
  & Trans.+ATAG+GATE (41+106) & \textbf{0.230} & \textbf{0.207} & 0.279 & \textbf{10.2} & \underline{0.059} & \textbf{0.057}  & \textbf{0.074} & \underline{12.6} & 0.256 & 0.341 & 0.380 \\ \hline \hline

\multirow{17}{*}{\shortstack{MIMIC \\ CXR}} & \multicolumn{9}{l}{\textbf{LSTM-based Model}} \\  \cline{2-13}
& WordSAT (47)~\cite{Xu2015Show} & 0.326 & 0.294  & 0.290 & 10.0 & 0.109 & 0.099  & 0.174 & 17.3 & 0.160 & 0.249 & \underline{0.082}\\
 & AdaAttn (47)~\cite{lu2017knowing} & \underline{0.367} & \underline{0.338} & \underline{0.334} & \underline{12.0}  & \underline{0.135} & \underline{0.130}  & 0.177 & 25.3 & 0.151 & 0.248 & \textbf{0.096} \\
 & SentSAT (47)~\cite{yuan2019automatic} & 0.366 & 0.329  & {0.326} & 11.3  & 0.122 & 0.121  & 0.186 & 20.0 & \textbf{0.182} & 0.252 & 0.073  \\
 & CoAttn (47)~\cite{jing2018automatic}  & 0.315 & 0.288  & 0.286 & 9.0  & 0.121 & 0.108  & 0.164 & 17.7 & \underline{0.181} & \textbf{0.253} & 0.070 \\
 & SentSAT (47+209)~\cite{yuan2019automatic} & 0.359 & 0.315  & 0.312 & \underline{12.0} & \textbf{0.139} & 0.131  & 0.181 & 22.0 & 0.178 & 0.247 & 0.065 \\ \cline{2-13}
 & SentSAT+AG (47) & 0.313 & 0.301 & \textbf{0.367} & 8.0 & 0.096 & 0.081 & 0.139 & 16.7 & 0.175 & \underline{0.250} & 0.068 \\
 & SentSAT+ATAG (47+209) & \underline{0.367} & 0.304 & 0.301 & \textbf{13.0} & 0.123 & 0.131 & \underline{0.188} & \underline{27.5} & \underline{0.181} & 0.249 & 0.080 \\
 & SentSAT+ATAG+GATE (47+209)  & \textbf{0.403} & \textbf{0.353}  & 0.291 & \textbf{13.0} & 0.125 & \textbf{0.135}  & \textbf{0.196} & \textbf{28.0} & 0.176 & \underline{0.250} & 0.079 \\ \cline{2-13}
& \multicolumn{9}{l}{\textbf{Transformer-based Model}} \\  \cline{2-13}
 & Transformer~\cite{vaswani2017attention} & 0.279 & 0.269 & 0.267 & 13.0 & 0.095 & 0.087 & 0.188 & 27.0 & 0.126 & 0.164 & \underline{0.167} \\
  & M2 Trans.~\cite{cornia2020meshed} & \underline{0.440} & \underline{0.391} & \underline{0.385} & \underline{13.3} & 0.140 & {0.135} & 0.231 & 35.7 & \textbf{0.159} & \textbf{0.250} & 0.100 \\
  & R2Gen~\cite{chen2020generating} & 0.268 & 0.298 & 0.293 & 13.0 & 0.098 & 0.104 & 0.159 & 26.0 & 0.124 & 0.160 & \textbf{0.170} \\
  & R2Gen-CMN~\cite{chen2021cross} & 0.313 & 0.309 & 0.305 & 10.0 & 0.119 & 0.108 & 0.182 & 26.0 & 0.123 & 0.163 & 0.128  \\ \cline{2-13}
  & Trans.+AG (47) & 0.400 & 0.371 & 0.367 & 12.7 & 0.150 & 0.152 & 0.215 & 33.7 & 0.142 & \underline{0.237} & 0.086 \\
  & Trans.+ATAG (47+209) & \textbf{0.442} & \textbf{0.400}  & \textbf{0.395} & \textbf{14.0} & \textbf{0.158} & \underline{0.164} & \underline{0.258} & \underline{40.0} & \underline{0.151} & {0.227} & 0.166\\
  & Trans.+ATAG+GATE(47+209) & 0.417 & 0.377 & 0.372 & \textbf{14.0} & \underline{0.149} & \textbf{0.172} & \textbf{0.266} & \textbf{41.0} & 0.145 & 0.225 & 0.160 \\ 

\hline
\end{tabular}

\end{table*}

To evaluate the effectiveness of generating clinically accurate reports, we adopt the \textit{clinical efficacy} metric and our proposed metric \texttt{RadRQI}.
The clinical efficacy score essentially measures the accuracy of 14 clinical observations\footnote{14 clincal observations includes: \textit{No finding}, \textit{Enlarged Cardiomediastinum}, \textit{Cardiomegaly}, \textit{Lung lesion}, \textit{Lung opacity}, \textit{Edema}, \textit{Consolidation}, \textit{Pneumonia}, \textit{Atelectasis}, \textit{Pneumothorax}, \textit{Pleural effusion}, \textit{Pleural other}, \textit{Fracture}, \textit{Support devices}} by comparing CheXpert-based labeling results obtained from the generated reports and the ground truth. The \texttt{CE} score reflects the correctness of the existence status of certain abnormalities and normality mentioned in the generated report. 
In Table \ref{gen:all}, we report the accuracy of 5 of 14 most represented observations, denoted as \texttt{CE(5)}\footnote{Five most represented observations includes: \textit{Atelectasis}, \textit{Cardiomegaly}, \textit{Consolidation}, \textit{Edema}, and \textit{Pleural effusion}.} ~\cite{miura2021improving}, 13 of 14 abnormality observations, denoted as \texttt{CE(14-1)}\footnote{Excluding \textit{No finding} from 14 clinical observations}~\cite{li2018hybrid} and 14 clinical observations, denoted as \texttt{CE(14)}~\cite{chen2020generating,chen2021cross,najdenkoska2021variational}. 

For models using LSTM-based decoders, we notice that \texttt{ATAG+GATE} can enhance them to generate more clinically accurate reports based on the \texttt{CE} metrics. 
It suggests that integrating our proposed ATAG and the gating mechanism with the LSTM decoder can enhance both accuracy and coverage of present abnormalities in the generated reports.

We also test models with Transformer-based decoders \texttt{Trans.+ATAG}. They obtain either the best or comparable scores of \texttt{CE(5)}, \texttt{CE(14-1)} and abnormality coverage \texttt{CE(Hits)}, indicating that a more powerful decoder can better utilize the ATAG embedding to enhance the clinical accuracy of the generated reports.
In particular, we notice that \texttt{Transformer} can achieve a high \texttt{CE(14)} score for IU XRay but not for \texttt{CE(14-1)} and \texttt{CE(Hits)} scores. This is due to the fact that
the model generates common sentences like ``No finding'' which make the accuracy of the ``normality'' high. Without taking into account ``normality'', \texttt{CE(14-1)} and \texttt{CE(Hits)} scores drop sharply, indicating that  many abnormalities are in fact missed. 

We also compare methods using the \texttt{RadRQI} score which corresponds to the accuracy based on a larger number of abnormalities and their associated attributes.
In general, the value of \texttt{RadRQI} score is lower than the \texttt{CE} score because the evaluation is more strict with the accuracy of the associated attributes also taken into consideration.
Our results again show that the ATAG-based models outperform the existing methods based on the \texttt{RadRQI} score.

This affirms the effectiveness of ATAG-based models to generate clinically accurate reports with more fine-grained details. 
Also, they can cover more abnormalities in the generated reports.
According to Table \ref{gen:all}, \texttt{Trans.+ATAG(+GATE)} is able to detect at least five more abnormalities than the evaluated baseline models. 
By further contrasting different decoders integrated with \texttt{ATAG}, those integrated with LSTM-based decoders perform better for the IU XRay dataset (smaller scale). 
Meanwhile, an average 61.9\% improvement
can be achieved by integrating \texttt{ATAG} with the Transformer-based decoder for the MIMIC CXR dataset. 

To evaluate the language quality of the generated reports, we adopt language quality metrics (e.g., BLEU), as reported in Table.~\ref{gen:all}. We notice that the gain in clinical accuracy due to the incorporation of the proposed ATAG and GATE does not compromise the language quality.
Noted that we used only the vanilla Transformer as the decoder in the experiment.
We anticipate that more powerful decoders (such as MemoryTrans.~\cite{chen2020generating} and AlignTrans.~\cite{you2021alignTransformer}), if adopted, should be able to further enhance the overall performance.

\begin{figure*}
 \centering
 \subfigure[LSTM-based Model in IU XRay]{
 \includegraphics[width=0.45\textwidth]{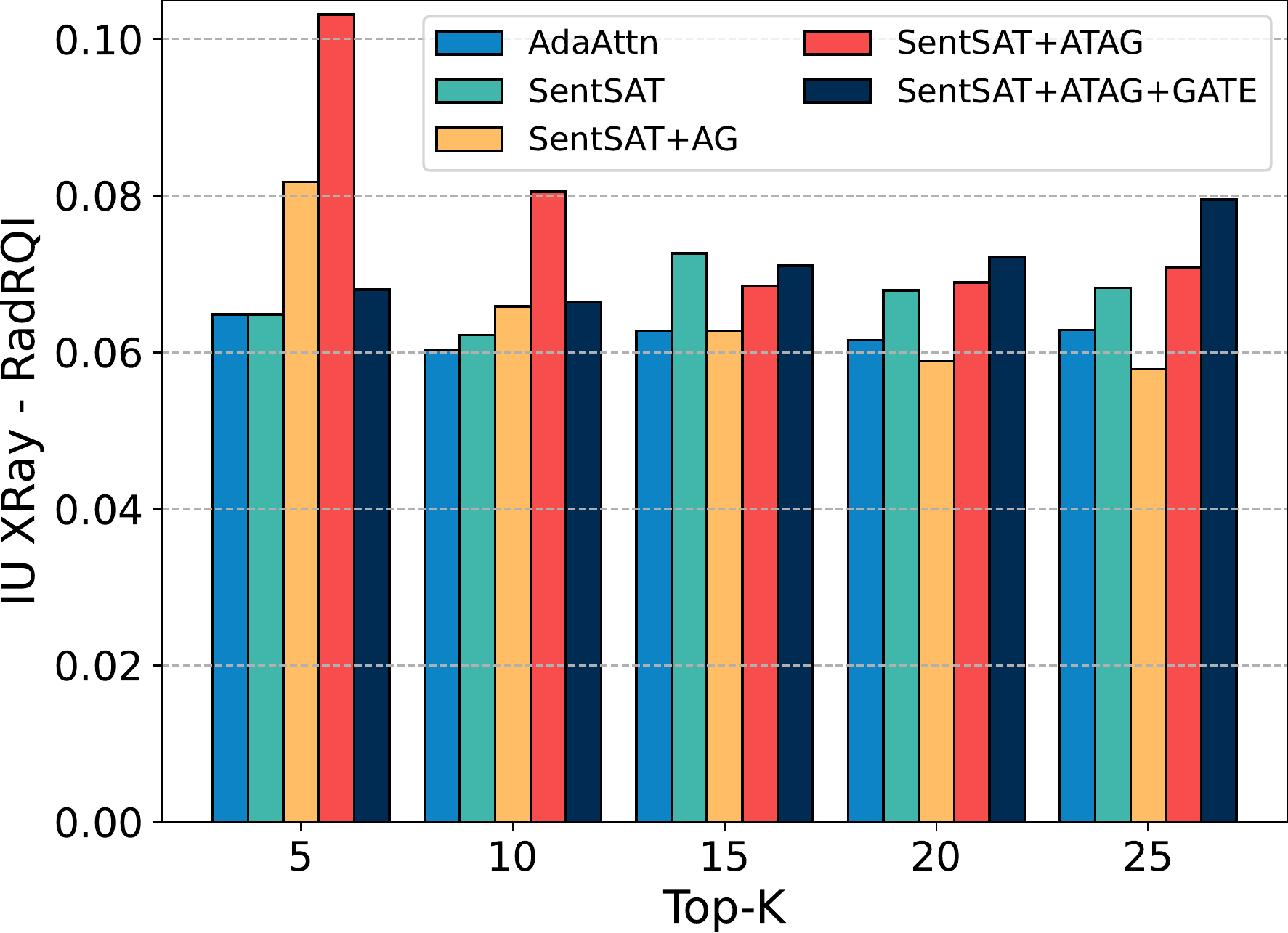}
 }
 \quad
 \subfigure[Transformer-based Model in IU XRay]{
 \includegraphics[width=0.45\textwidth]{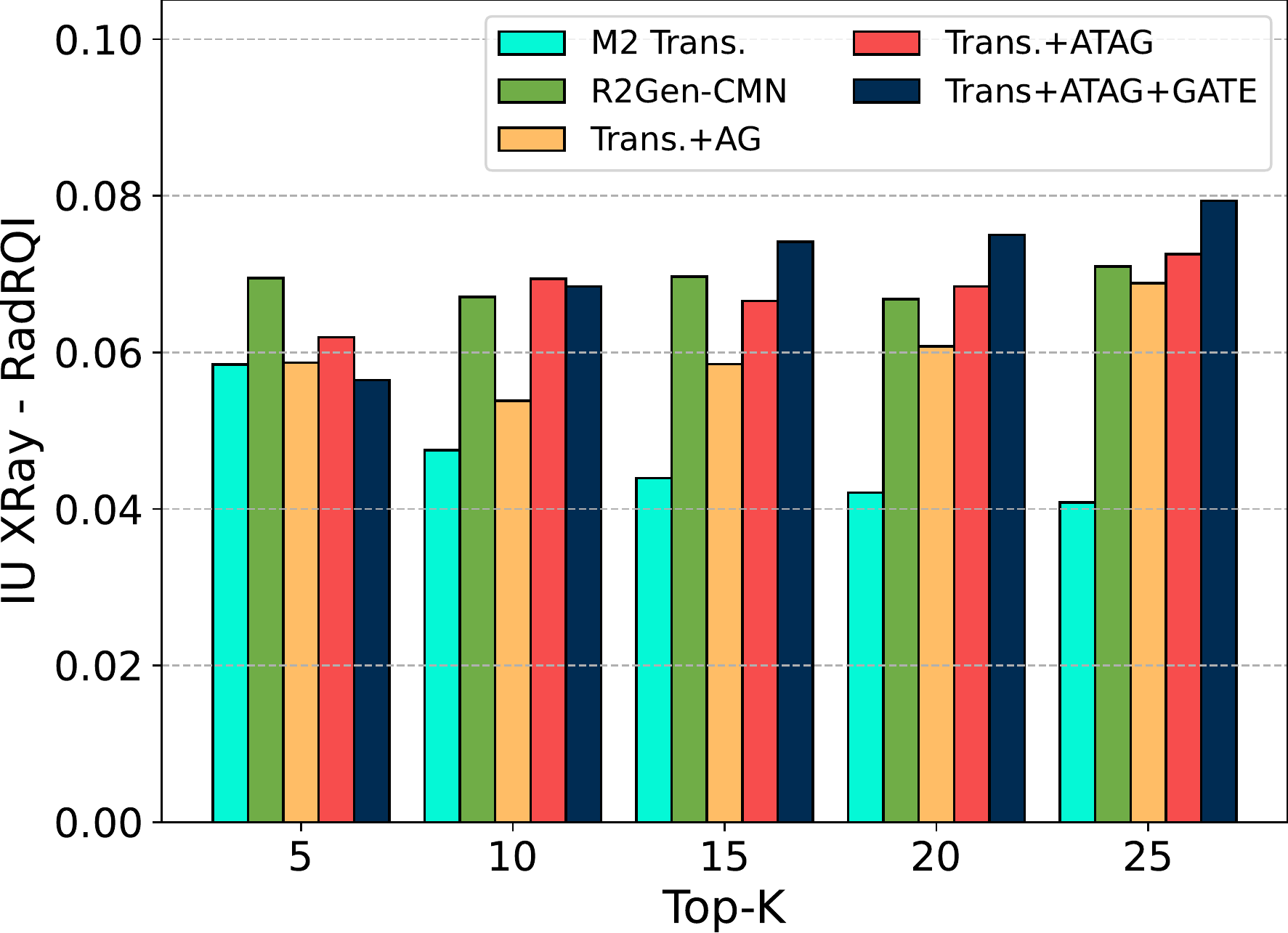}
 }
 \quad
 \subfigure[LSTM-based Model in MIMIC CXR]{
 \includegraphics[width=0.45\textwidth]{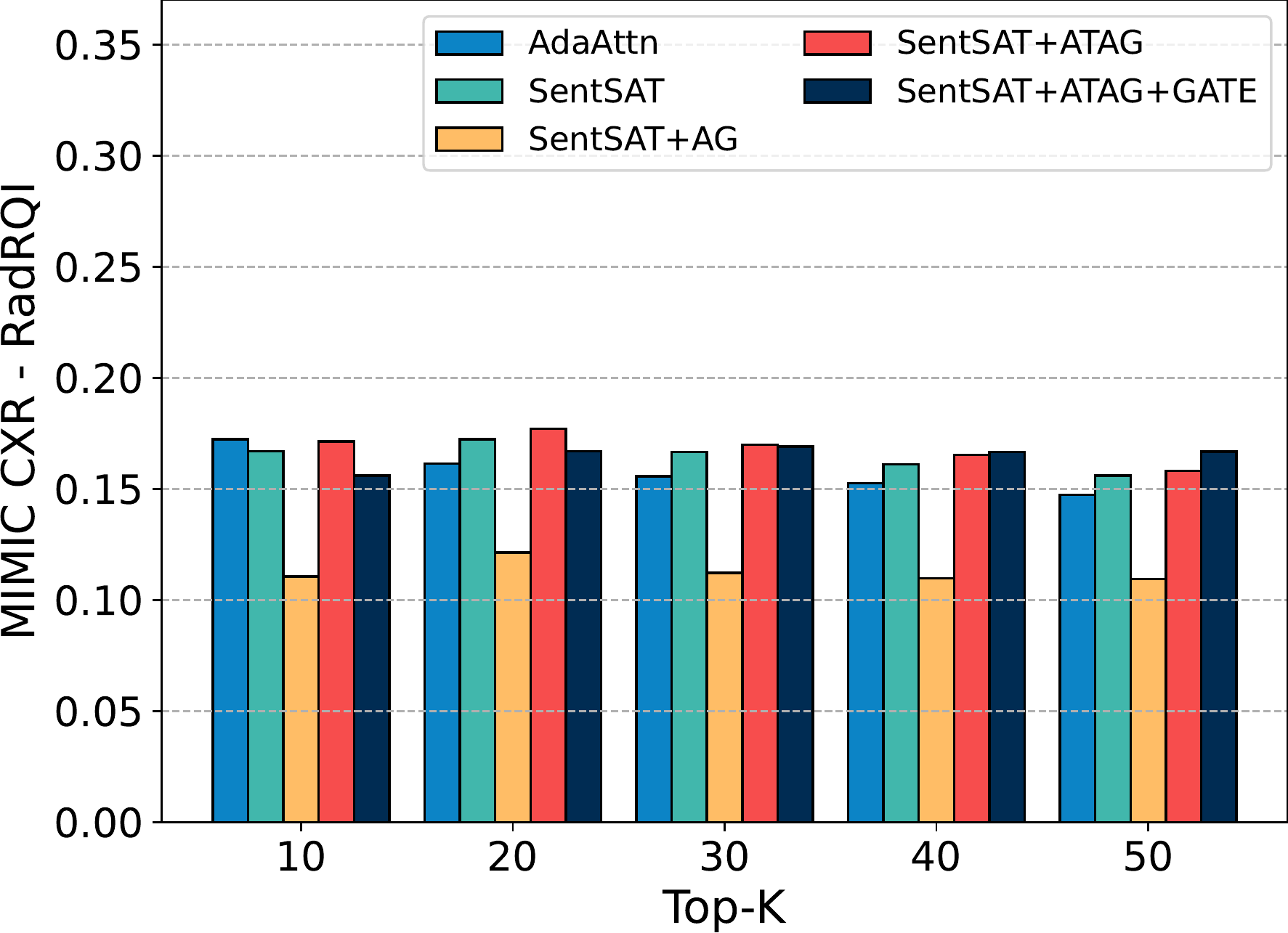}
 }
 \quad
 \subfigure[Transformer-based Model in  MIMIC CXR]{
 \includegraphics[width=0.45\textwidth]{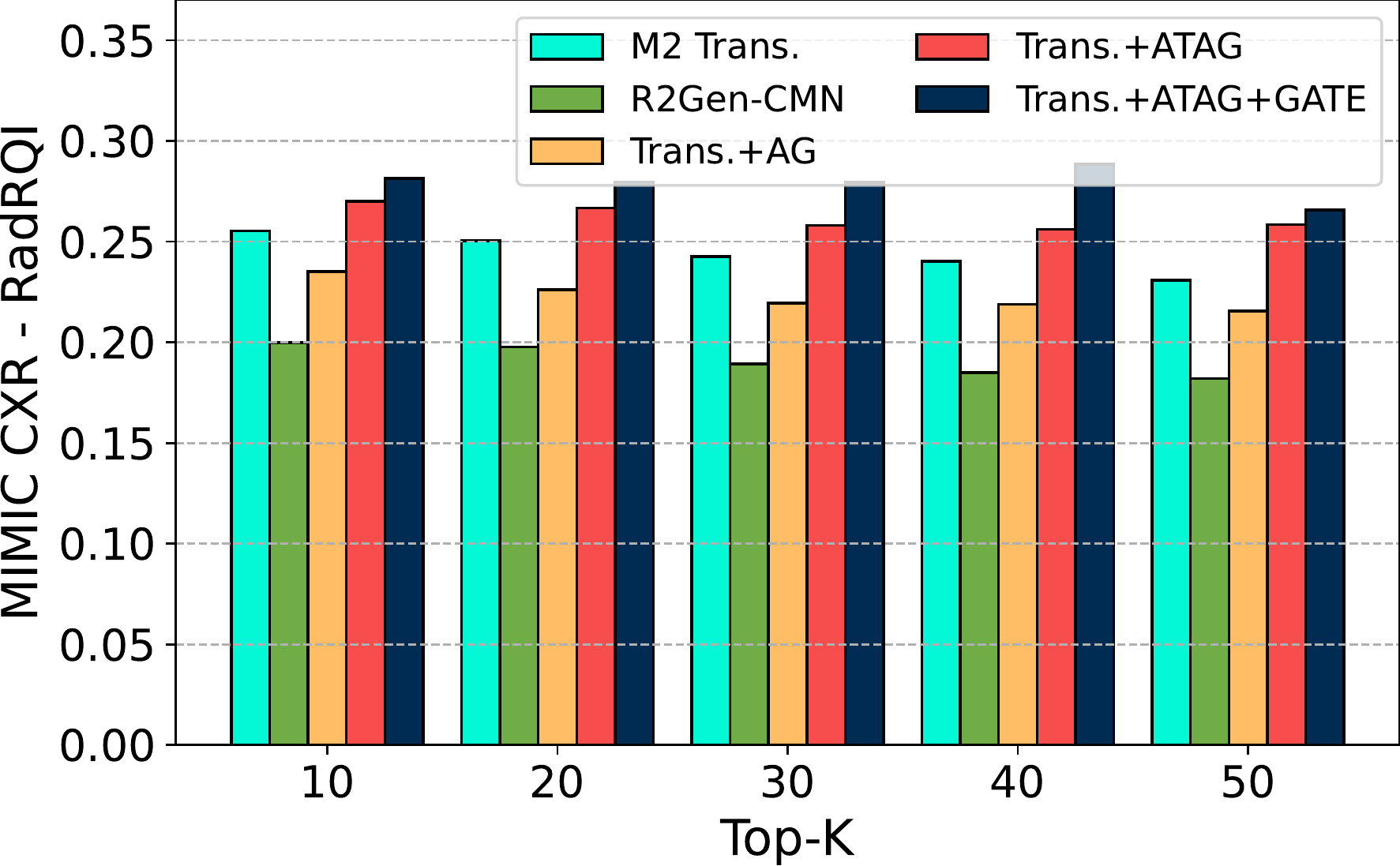}
 }
 \caption{Performance of the baselines and our proposed approach with respect to various settings of Top-K in calculating the RadRQI score.}
 \label{fig:my_label}
\end{figure*}

\subsection{Sensitivity Analysis}

\subsubsection{Size of the ATAG}
To analyze how the size of the introduced ATAG affects the clinical accuracy of the generated reports, we construct ATAGs with different numbers of abnormalities and attributes. To do that, we put different thresholds on the occurrence frequency of the abnormalities and attributes in the dataset,
as shown in Table.~\ref{tab:abnatr}. Table.~\ref{tab:gensize} shows the performance of the models with ATAGs of different size.

In general, the ATAG of larger size can cover more abnormalities so that the decoder has a higher chance to generate descriptions with more abnormalities and attributes. 
With reference to the \texttt{RadRQI(14-1)} and \texttt{RadRQI(TopK)} columns, the larger ATAG can improve the accuracy of certain and overall abnormalities with associated attributes. 
One possible reason could be that models with more detailed abnormalities and their relationships captured with the help of ATAGs can allow them to distinguish better different abnormalities. 

For the smaller dataset IU XRay, the improvement from \texttt{ATAG(28+79)} to \texttt{ATAG(41+106)} is marginal which is probably due to the fact that the additional abnormalities and attributes are rare and capturing them as well does not contribute much to the overall accuracy. When the dataset is large (like MIMIC CXR), the improvement gain due to increase in the ATAG size is obvious. E.g.,  
Transformer-based decoder with \texttt{ATAG (47+209)} can achieve an improvement of 15.0\% on \texttt{RadRQI(TopK)}.
This observation is important because the setting is closer to the real-world situations. 
We also observe that the adoption of LSTM-based decoders could limit the effectiveness of ATAG for large datasets. For instance, the highest \texttt{RadRQI(TopK)} scores of \texttt{SentSAT+ATAG} is obtained by \texttt{ATAG(35+165)} and larger ATAGs in fact decrease the performance in our experiment. It also indicates the optimal design of ATAG will depend on the decoder integrated.

\begin{table*}[!t]
\centering
\caption{Performance comparison of report generation models with different sizes of ATAG on two clinical accuracy metrics. The best scores are in bold face and the second best are underlined. ``Top-K'' are set to top-25 and top-50 in IU XRay and MIMIC CXR, respectively.}
\label{tab:gensize}
\begin{tabular}{c|c|l|cccc|cccc}
\hline
\multirow{2}{*}{\textbf{Dataset}} & \multirow{2}{*}{\textbf{(Abn.\# + Atr.\#)}} & \multirow{2}{*}{\textbf{Model}} & \multicolumn{4}{|c|}{\textbf{Clinical Efficacy}} & \multicolumn{4}{|c}{\textbf{RadRQI}} \\
& & & \textbf{(5)} & \textbf{(14-1)} & \textbf{(14)} & \textbf{Hits} & \textbf{(5)} & \textbf{(14-1)} & \textbf{Top-K} & \textbf{Hits} \\
\hline
\multirow{18}{*}{\shortstack{IU \\ XRay}} 
& \multicolumn{10}{l}{\textbf{LSTM-based Model}} \\ \cline{2-11}
& \multirow{3}{*}{(41+106)} & SentSAT & 0.157 & 0.123 & 0.229 & 5.6 & 0.052 & 0.048 & 0.056 & 6.4 \\
& & SentSAT+ATAG & 0.190 & 0.145 & \underline{0.244} & 7.2 & \underline{0.062} & \underline{0.065} & 0.069 & 10.2 \\
& & SentSAT+ATAG+GATE & 0.216 & 0.178 & \textbf{0.263} & 8.6 & \textbf{0.066} & \textbf{0.068} & \textbf{0.079} & \textbf{12.4} \\
\cline{2-11}

& \multirow{3}{*}{(28+79)} & SentSAT & 0.160 & 0.114 & 0.224 & 5.2 & 0.043 & 0.042 & 0.052 & 7.2 \\
 & & SentSAT+ATAG & 0.201 & 0.164 & 0.223 & 8.8 & 0.055 & 0.058 & 0.060 & 9.2 \\
 & & SentSAT+ATAG+GATE & 0.208 & 0.166 & \underline{0.244} & 8.4 & 0.049 & 0.050 & \underline{0.074} & 10.6 \\
\cline{2-11}

& \multirow{3}{*}{(23+64)} & SentSAT & 0.209 & 0.168 & 0.230 & \underline{10.2} & 0.036 & 0.040 & 0.060 & 10.0 \\
 & & SentSAT+ATAG & \underline{0.225} & \underline{0.182} & 0.239 & 9.2 & 0.048 & 0.054 & 0.067 & 10.4 \\
 & & SentSAT+ATAG+GATE & \textbf{0.226} & \textbf{0.191} & 0.234 & \textbf{10.8} & 0.039 & 0.043 & 0.065 & \underline{11.0} \\
\cline{2-11}

& \multicolumn{10}{l}{\textbf{Transformer-based Model}} \\ \cline{2-11}
& - & Transformer & 0.124 & 0.112 & \textbf{0.310} & 5.0 & \underline{0.052} & 0.038 & 0.072 & 9.0 \\
\cline{2-11}

 & \multirow{2}{*}{(41+106)} & Trans.+ATAG & 0.184 & 0.178 & 0.262 & 10.0 & 0.050 & 0.047 & 0.072 & \textbf{13.8} \\
 & & Trans.+ATAG+GATE & \textbf{0.230} & \textbf{0.207} & \underline{0.279} & 10.2 & \textbf{0.059} & \textbf{0.057} & \underline{0.074} & 12.6 \\
\cline{2-11}

& \multirow{2}{*}{(28+79)} & Trans.+ATAG & 0.198 & 0.188 & 0.267 & \textbf{10.8} & 0.045 & \underline{0.053} & \textbf{0.075} & \textbf{13.8} \\
 & & Trans.+ATAG+GATE & \underline{0.208} & \underline{0.189} & 0.271 & 10.2 & 0.049 & 0.051 & 0.074 & \underline{13.3} \\
\cline{2-11}

& \multirow{2}{*}{(23+64)} & Trans.+ATAG & 0.166 & 0.160 & 0.251 & 9.8 & 0.043 & 0.045 & 0.069 & 12.6 \\
 & & Trans.+ATAG+GATE & 0.200 & 0.174 & 0.256 & \underline{10.4} & 0.037 & 0.041 & 0.064 & 11.6 \\
\hline \hline 
\multirow{18}{*}{\shortstack{MIMIC \\ CXR}}  & \multicolumn{10}{l}{\textbf{LSTM-based Model}} \\ \cline{2-11}
& \multirow{3}{*}{(47+209)} & SentSAT & 0.359 & 0.315 & 0.312 & 12.0 & \textbf{0.139} & 0.131 & 0.181 & 22.0 \\
 & & SentSAT+ATAG & 0.367 & 0.304 & 0.301 & \underline{13.0} & 0.123 & 0.131 & 0.188 & \underline{27.5} \\
 & & SentSAT+ATAG+GATE & \textbf{0.403} & \textbf{0.353} & 0.251 & \underline{13.0} & 0.125 & 0.135 & 0.196 & \textbf{28.0} \\
\cline{2-11}

& \multirow{3}{*}{(35+165)} & SentSAT & 0.253 & 0.291 & 0.286 & \underline{13.0} & \underline{0.136} & 0.133 & 0.202 & 20.0 \\
 & & SentSAT+ATAG & 0.323 & 0.267 & 0.265 & 10.0 & 0.135 & \textbf{0.149} & \textbf{0.227} & 23.3 \\
 & & SentSAT+ATAG+GATE & \underline{0.378} & \underline{0.334} & \textbf{0.331} & \underline{13.0} & 0.135 & \underline{0.141} & \underline{0.223} & 24.6 \\
\cline{2-11}

 & \multirow{3}{*}{(26+129)} & SentSAT & 0.288 & 0.293 & 0.288 & \underline{13.0} & 0.122 & 0.120 & 0.170 & 21.0 \\
 & & SentSAT+ATAG & 0.292 & 0.333 & \underline{0.328} & \underline{13.0} & 0.110 & 0.136 & 0.189 & 27.0 \\
 & & SentSAT+ATAG+GATE & 0.277 & 0.304 & 0.327 & \textbf{14.0} & 0.116 & 0.130 & 0.191 & 27.0 \\
\cline{2-11}

& \multicolumn{10}{l}{\textbf{Transformer-based Model}} \\ \cline{2-11}
 & - & Transformer & 0.279 & 0.269 & 0.267 & \underline{13.0} & 0.095 & 0.087 & 0.188 & 30.0 \\
\cline{2-11}

& \multirow{2}{*}{(47+209)} & Trans.+ATAG & \underline{0.419} & 0.368 & 0.363 & \textbf{14.0} & \textbf{0.164} & \underline{0.166} & \underline{0.228} & 36.0 \\
& & Trans.+ATAG+GATE & 0.417 & 0.377 & 0.372 & \textbf{14.0} & 0.149 & \textbf{0.172} & \textbf{0.266} & \textbf{41.0} \\
\cline{2-11}

& \multirow{2}{*}{(35+165)} & Trans.+ATAG & 0.408 & \textbf{0.411} & \textbf{0.400} & \textbf{14.0} & 0.140 & 0.151 & 0.235 & 34.0 \\
 & & Trans.+ATAG+GATE & \textbf{0.422} & \underline{0.393} & \underline{0.387} & \underline{13.0} & 0.149 & 0.149 & 0.243 & 35.5 \\
\cline{2-11}

& \multirow{2}{*}{(26+129)} & Trans.+ATAG & 0.400 & 0.372 & 0.368 & \textbf{14.0} & 0.139 & 0.147 & 0.219 & \underline{37.0} \\
 & & Trans.+ATAG+GATE & 0.410 & 0.374 & 0.369 & 12.0 & \underline{0.153} & 0.153 & 0.221 & 36.0 \\
\hline
\end{tabular}
\end{table*}

\subsubsection{Dataset-Specific Abnormalities and Attributes}
\begin{figure*}
 \centering
 \subfigure[IU XRay dataset.]{
 \includegraphics[width=0.95\textwidth]{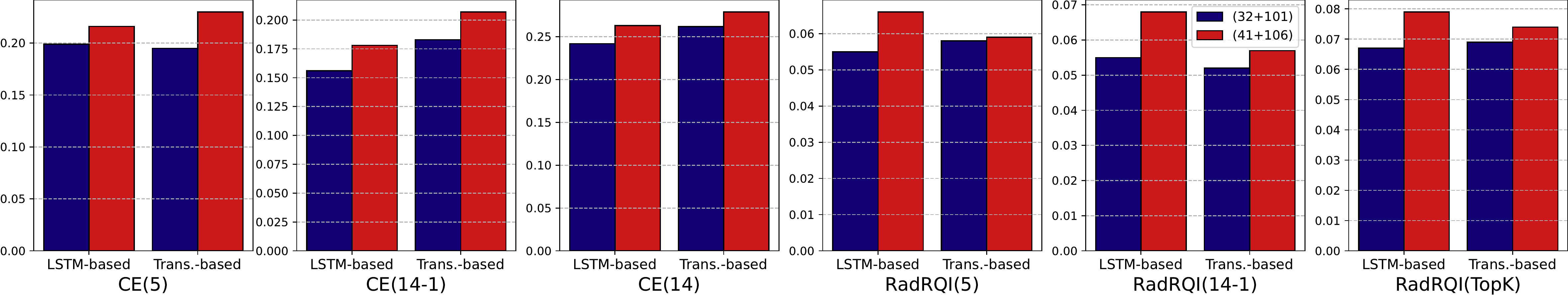}
 }
 \quad
 \subfigure[MIMIC CXR dataset.]{
 \includegraphics[width=0.95\textwidth]{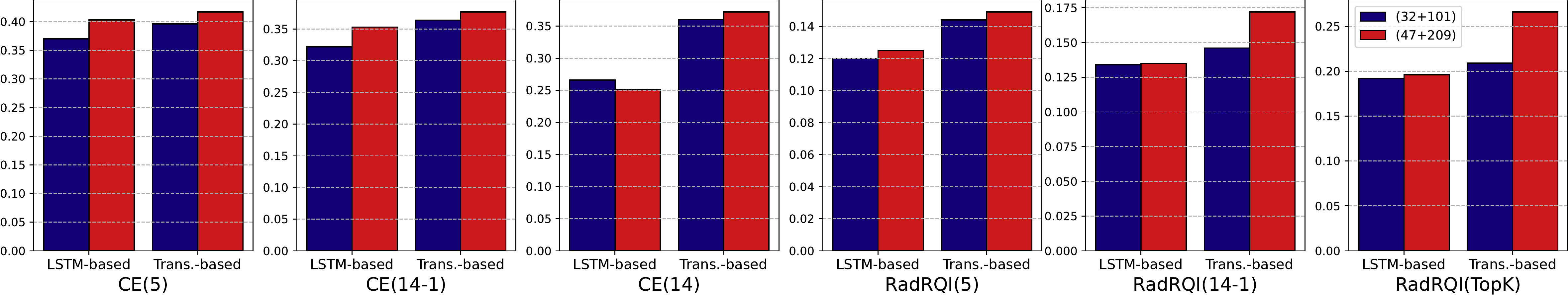}
 }
 \caption{Performance of \texttt{SentSAT+ATAG+GATE} and \texttt{Trans.+ATAG+GATE} based on a common set of anormalities or the set of anormalities specific to two different datasets.}
 \label{fig:data-specific}
\end{figure*}

Different datasets have their own sets of abnormalities and attributes where some are common and some specific to the corresponding dataset.
To show the importance of taking care of dataset-specific abnormalities and attributes, we construct \texttt{ATAG(32+101)} which is composed of 32 abnormalities and 101 attributes shared between \texttt{ATAG(41+106)} learned for IU XRay and \texttt{ATAG(47+209)} learned for MIMIC CXR. We compare the performance of the generic model and the dataset-specific ones, as shown in Fig.~\ref{fig:data-specific}.

With reference to the generic model \texttt{ATAG(32+101)}, 
22.8\% and 3.3\% improvement on \texttt{RadRQI(TopK)} can be obtained by modeling the specific abnormalities and attributes based on \texttt{SentSAT+ATAG(+GATE)}. For \texttt{Trans.+ATAG(+GATE)}, improvement of 9.9\% and 29.5\% can be achieved.
Modeling the dataset-specific abnormalities and attributes also improves the accuracy of certain abnormalities. We observe 15.2\% and 6.1\% improvement on average on \texttt{CE(14-1)} by integrating \texttt{ATAG(+GATE)} with LSTM- and Transformer-based decoders. 

Noted that, limited improvements are observed for \texttt{RadRQI(14-1)} scores. The possible explanation could be that the number of dataset-specific attributes of certain abnormalities is small, thus the related generated reports may show seldom different which makes \texttt{RadRQI(14-1)} scores similar.

\subsection{Case Study}

\begin{figure*}[!h]
\centering
\includegraphics[width=1\textwidth]{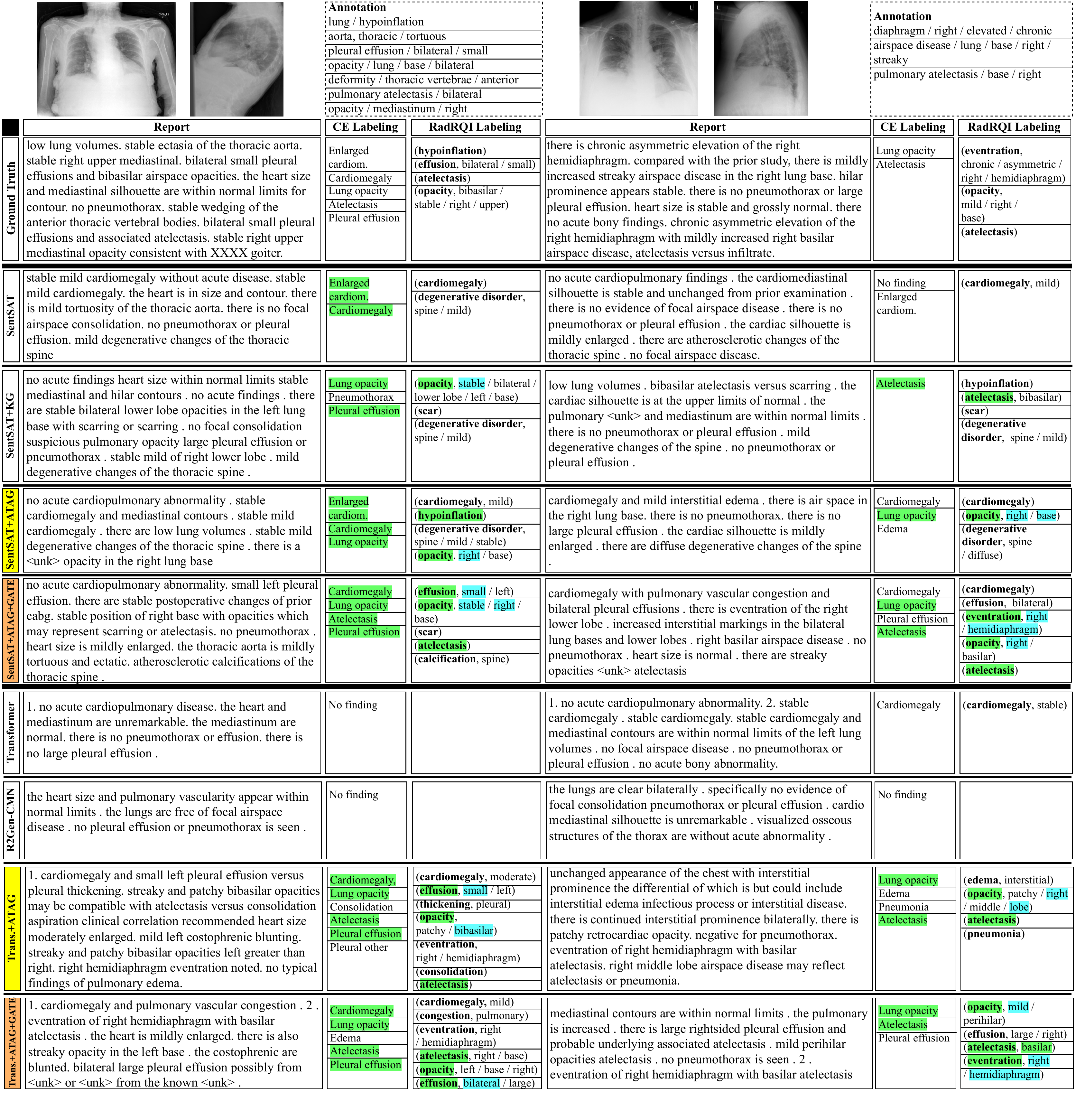}
\caption{Illustration of reports generated by the baseline model and models integrated with \texttt{ATAG} (yellow background color) and \texttt{GATE} (orange background color) on the IU XRay dataset. The first section (1st row) is the ground truth reports, second section (2nd-5th rows) is the generated report by LSTM-based models and third section (6th-9th rows) is the generated report by Transformer-based models. The correct abnormality and attribute terms are highlighted with green and blue colors. The correct term The expert-labeled annotations provided by \cite{demner2016preparing} are also attached.}
\label{fig:case1}
\end{figure*}

Fig.~\ref{fig:case1} shows two cases from the IU X-Ray dataset.
For each case, we visualize the ground truth report and the reports generated by baseline models and the proposed ATAG-based models.

We apply a post-processing step of removing duplicated or short sentences from the generated reports, and show the disease/abnormality keywords extracted by CheXpert labeling toolkit used by \texttt{CE} metrics and RadLex+RadGraph used by \texttt{RadRQI} from the ground truth and the generated reports accordingly. 

As illustrated, we observe that integrating ATAG can generate more accurate abnormalities, while the gating mechanism is able to further increase the accuracy of associated attributes for the mentioned abnormality. 
The comparison also suggests that utilizing the attributed abnormality embedding is able to facilitate detecting the correct abnormalities and associated attributes.

Yet, it is also observed that some abnormalities cannot be well distinguished due to several reasons. For example, in Fig.~\ref{fig:case1} 1st case, \texttt{SentSAT+ATAG}, \texttt{Trans+ATAG} and \texttt{Trans.+ATAG+GATE} are reported to detect the ``Cardiomegaly'', which could be caused by the denominated white regions shown in the center X-ray image which makes models hard to distinguish the heart outline and the below region.
Also, when the visual pattern of ``atelectasis'' is detected, the \texttt{Trans.+ATAG+GATE} generates ``\textit{less severe consolidation in the right lower lobe is either pneumonia or atelectasis}'' which mentions three possible abnormality observations \textit{consolidation}, \textit{pneumonia} and \textit{atelectasis} that have similar patterns. It suggests that the generation model attempts to point out more potential abnormalities so that the present abnormal observations will not be missed as far as possible. 

\section{Conclusion}
In this paper, we propose to automatically construct a fine-grained attributed abnormality graph (ATAG) and the corresponding embedding for representing abnormalities in X-ray images. 
In particular, an ATAG with an abnormality graph of which each node is paired with a specific attribute graph. To the best of our knowledge, this is the first attempt to construct the detailed graph structure and then the embedding automatically from annotated reports. 

A hierarchical attention mechanism is proposed to aggregate the abnormality and attribute embeddings, 
and a gate mechanism is employed to integrate ATAG embedding into both LSTM- and Transformer-based decoder for radiology report generation.
We performed comprehensive empirical evaluation on the benchmark datasets. Our experiment results show that the proposed ATAG-based deep model can improve the accuracy for both abnormality classification and radiology report generation compared to the state-of-the-art models. Potential future directions include consideration of ambiguous and potentially incorrect annotations, as well as integration of EHR data of different modalities to further achieve clinical accuracy. 

\bibliographystyle{IEEEtran}
\bibliography{ref}




%







\end{document}